\documentclass[reprint,superscriptaddress,aps,prx]{revtex4-2}

\usepackage[usenames,dvipsnames]{xcolor}
\usepackage[utf8]{inputenc}
\usepackage{bm}
\usepackage{amsfonts}
\usepackage{graphicx,psfrag}

\usepackage{booktabs}
\usepackage{amsmath}
\usepackage{calc}
\usepackage{makecell}

\usepackage{nicefrac}

\usepackage{enumitem}
\usepackage{float}
\usepackage{natbib}
\usepackage{tikz}
\usepackage{soul,xcolor}
\usepackage[colorlinks=true,linkcolor=Blue,citecolor=Blue,urlcolor=Blue]{hyperref}

\usepackage{xfp}

\usepackage[caption=false]{subfig}
\usepackage[capitalise]{cleveref}

\newcommand{\x}{\mathbf{x}}
\newcommand{\f}{\mathbf{f}}
\newcommand{\g}{\mathbf{g}}
\newcommand{\F}{\mathbf{F}}
\newcommand{\G}{\mathbf{G}}
\newcommand{\y}{\mathbf{y}}
\newcommand{\res}{\mathbf{r}}
\newcommand{\W}{\mathbf{W}}
\newcommand{\Wr}{\W_r}
\newcommand{\Win}{\W_\text{in}}
\newcommand{\bias}{\mathbf{b}}
\newcommand{\Y}{\mathbf{Y}}
\newcommand{\xt}{\tilde{x}}
\newcommand{\yt}{\tilde{y}}
\def\*#1{\mathbf{#1}}

\newcommand{\Ntrain}{N_\textnormal{train}}
\newcommand{\Ntraj}{N_\textnormal{traj}}
\newcommand{\NRBF}{N_\textnormal{RBF}}
\newcommand{\mlin}{m_\textnormal{lin}}
\newcommand{\mnonlin}{m_\textnormal{nonlin}}

\def\multiset#1#2{\ensuremath{\left(\kern-.3em\left(\genfrac{}{}{0pt}{}{#1}{#2}\right)\kern-.3em\right)}}

\newlength{\depthofsumsign}
\newlength{\totalheightofsumsign}
\newlength{\heightanddepthofargument}

\newcommand{\nsum}[1][1.4]{
    \mathop{%
        \raisebox
            {-#1\depthofsumsign+1\depthofsumsign}
            {\scalebox
                {#1}
                {$\displaystyle\sum$}%
            }
    }
}

\makeatletter
\newcommand*{\DivideLengths}[2]{%
  \strip@pt\dimexpr\number\numexpr\number\dimexpr#1\relax*65536/\number\dimexpr#2\relax\relax sp\relax
}
\makeatother

\begin{document}

\title{Catch-22s of Reservoir Computing}

\author{Yuanzhao Zhang}
\affiliation{Santa Fe Institute, 1399 Hyde Park Road, Santa Fe, NM 87501, USA}

\author{Sean P. Cornelius}
\affiliation{Department of Physics, Toronto Metropolitan University, Toronto, ON, M5B 2K3, Canada}

\begin{abstract} 
    Reservoir Computing (RC) is a simple and efficient model-free framework for forecasting the behavior of nonlinear dynamical systems from data. Here, we show that there exist commonly-studied systems for which leading RC frameworks struggle to learn the dynamics unless key information about the underlying system is already known. We focus on the important problem of basin prediction---determining which attractor a system will converge to from its initial conditions. First, we show that the predictions of standard RC models (echo state networks) depend critically on warm-up time, requiring a warm-up trajectory containing almost the entire transient in order to identify the correct attractor. Accordingly, we turn to Next-Generation Reservoir Computing (NGRC), an attractive variant of RC that requires negligible warm-up time. By incorporating the exact nonlinearities in the original equations, we show that NGRC can accurately reconstruct intricate and high-dimensional basins of attraction, even with sparse training data (e.g., a single transient trajectory). Yet, a tiny uncertainty in the exact nonlinearity can render prediction accuracy no better than chance. Our results highlight the challenges faced by data-driven methods in learning the dynamics of multistable systems and suggest potential avenues to make these approaches more robust.
\end{abstract}

\maketitle

\section{Introduction}

Reservoir Computing (RC) \cite{maass2002real,jaeger2004harnessing,lukovsevivcius2009reservoir,appeltant2011information,canaday2018rapid,carroll2018using,vlachas2020backpropagation,rafayelyan2020large,fan2020long,gottwald2021combining,zhong2021dynamic,nakajima2021reservoir} is a machine learning framework for time-series predictions based on recurrent neural networks. 
Because only the output layer needs to be modified, RC is extremely efficient to train.
Despite its simplicity, recent studies have shown that RC can be extremely powerful when it comes to learning unknown dynamical systems from data \cite{pathak2018model}.
Specifically, RC has been used to reconstruct attractors \cite{lu2018attractor,grigoryeva2023learning}, calculate Lyapunov exponents \cite{pathak2017using}, infer bifurcation diagrams \cite{kim2021teaching}, and even predict the basins of unseen attractors \cite{rohm2021model,roy2022model}.
These advances open the possibilities of using RC to improve climate modeling \cite{arcomano2022hybrid}, create digital twins \cite{antonik2018using}, anticipate synchronization \cite{weng2019synchronization,fan2021anticipating}, predict tipping points \cite{kong2021machine,patel2023using}, and infer network connections \cite{banerjee2021machine}.

Since the landmark paper demonstrating RC's ability to predict spatiotemporally chaotic systems from data \cite{pathak2018model}, there has been a flurry of efforts to understand the success as well as identify limitations of RC \cite{carroll2019network,jiang2019model,gonon2019reservoir,griffith2019forecasting,carroll2020reservoir,pyle2021domain,hart2021echo,platt2021robust,flynn2021multifunctionality,carroll2022optimizing}.
As a result, more sophisticated architectures have been developed to extend the capability of the original framework, such as hybrid \cite{pathak2018hybrid}, parallel \cite{wikner2020combining,srinivasan2022parallel}, and symmetry-aware \cite{barbosa2021symmetry} RC schemes.

One particularly promising variant of RC was proposed in 2021 and named Next Generation Reservoir Computing (NGRC) \cite{gauthier2021next}.
There, instead of having a nonlinear reservoir and a linear output layer, one has a linear reservoir and a nonlinear output layer \cite{bollt2021explaining}.
These differences, though subtle, confer several advantages: First, NGRC requires no random matrices and thus has much fewer hyperparameters that need to be optimized.
Moreover, each NGRC prediction needs exceedingly few data points to initiate (as opposed to thousands of data points in standard RC), which is especially useful when predicting the basins of attraction in multistable dynamical system \cite{gauthier2022learning}.

Understanding the basin structure is of fundamental importance for dynamical systems with multiple attractors.
Such systems include neural networks \cite{hopfield1982neural,li2018visualizing}, gene regulatory networks \cite{teschendorff2021statistical,rand2021geometry}, differentiating cells \cite{schiebinger2019optimal,saez2022statistically}, and power grids \cite{menck2013basin,menck2014dead}.
Basins of attraction provide a mapping from initial conditions to attractors and, in the face of noise or perturbations, tell us the robustness of each stable state.
Despite their importance, basins have not been well studied from a machine learning perspective, with most methods for data-driven modeling of dynamical systems currently focusing on systems with a single attractor.

In this Article, we show that the success of standard RC in predicting the dynamics of multistable systems can depend critically on having access to long initialization trajectories, while the performance of NGRC can be extremely sensitive to the choice of readout nonlinearity.
It has been observed that, for each new initial condition, a standard RC model needs to be ``warmed up'' with thousands of data points before it can start making predictions \cite{gauthier2022learning}.
In practice, such data will not exist for most initial conditions.
Even when they do exist, we demonstrate that the warm-up time series would often have already approached the attractor, rendering predictions unnecessary \footnote{Note that the warm-up time series is different from the training data and is only used after training has been completed.}.
In contrast, NGRC can easily reproduce highly intermingled and high-dimensional basins with minimal warm-up, provided the exact nonlinearity in the underlying equations is known.
However, a $1\%$ uncertainty on that nonlinearity can already make the NGRC basin predictions barely outperform random guesses.
Given this extreme sensitivity, even if one had partial (but imprecise) knowledge of the underlying system, a hybrid scheme combining NGRC and such knowledge would still struggle in making reliable predictions.

The rest of the paper is organized as follows. 
In \cref{sec:pendulum}, we introduce the first model system under study---the magnetic pendulum, which is representative of the difficulties of basin prediction in real nonlinear systems. 
In Sections \ref{sec:standard-rc-implementation}-\ref{sec:standard-rc-results}, we apply standard RC to this system, showing that accurate predictions rely heavily on the length of the warm-up trajectory.
We thus turn to Next Generation Reservoir Computing, giving a brief overview of its implementation in \cref{sec:ngrc-implementation}. 
We present our main results in \cref{sec:nonlinearity_pendulum}, where we characterize the effect of readout nonlinearity on NGRC's ability to predict the basins of the magnetic pendulum. 
We further support our findings using coupled Kuramoto oscillators in \cref{sec:kuramoto}, which can have a large number of coexisting high-dimensional basins. 
Finally, we discuss the implications of our results and suggest avenues for future research in \cref{sec:end}.

\section{The magnetic pendulum}
\label{sec:pendulum}

For concreteness, we focus on the magnetic pendulum \cite{motter2013doubly} as a representative model. It is mechanistically simple---being low-dimensional and generated by simple physical laws---and yet captures all characteristics of the basin prediction problem in general: the system is multistable and predicting which attractor a given initial condition will go to is nontrivial.

\begin{figure}[tb]
\centering
\includegraphics[width=.99\columnwidth]{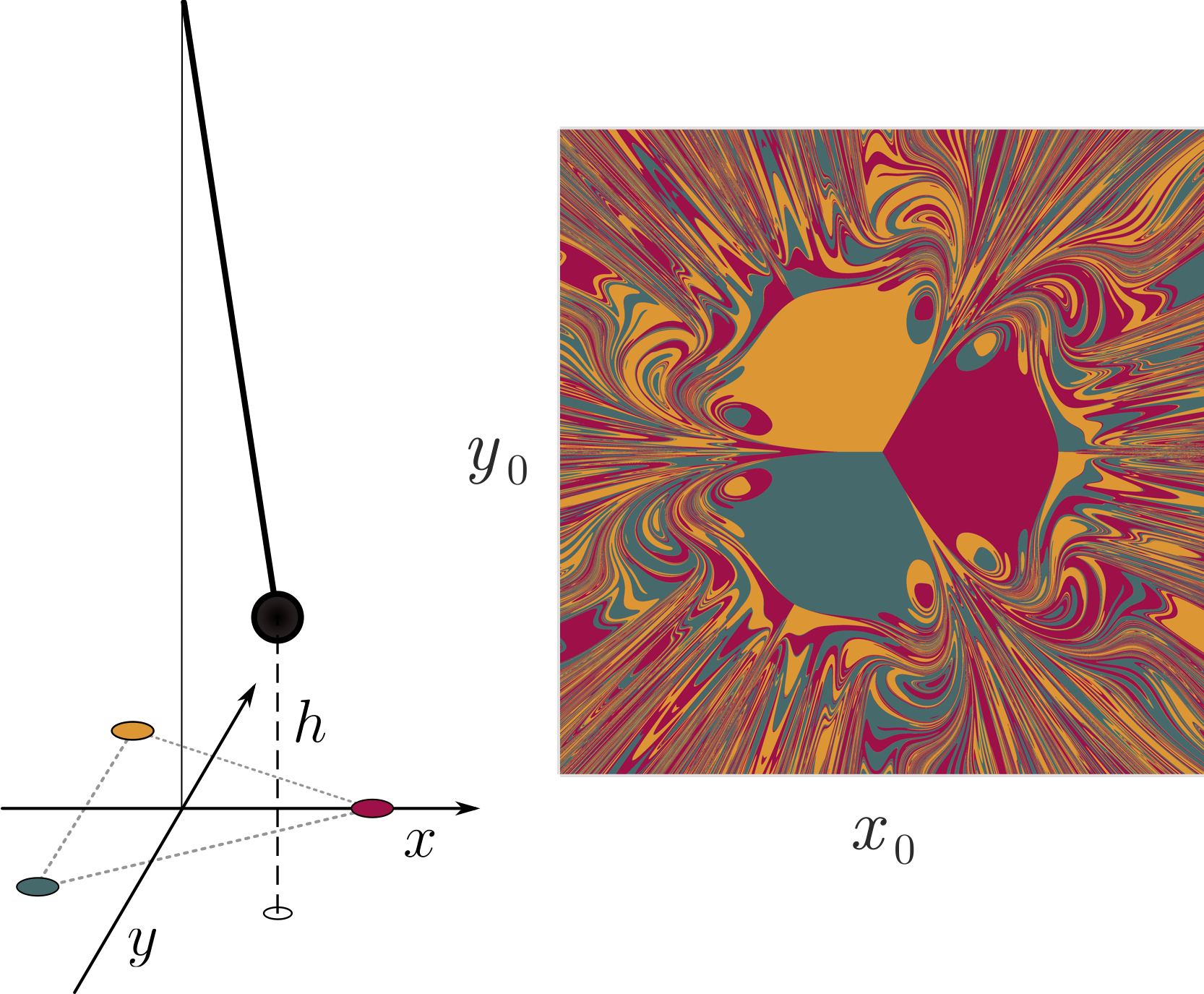}
\caption{
\textbf{Magnetic pendulum with three fixed-point attractors and the corresponding basins of attraction}. (Left) Illustration of the magnetic pendulum system. Three magnets are placed on a flat surface, each drawn in the color we use to denote the corresponding basin of attraction. The hollow circle indicates the $(x, y)$ coordinates of the pendulum bob, which together with the velocity  $(\dot{x}, \dot{y})$ fully specify the system's state. (Right) Basins of attraction for the region of initial conditions under study, namely states of zero initial velocity with $-1.5 \le x_0, y_0 \le 1.5$.}
\label{fig:schematic}
\end{figure}

The system consists of an iron bob suspended by a massless rod above three identical magnets, located at the vertices of an equilateral triangle in the $(x, y)$ plane (\cref{fig:schematic}). The bob moves under the influence of gravity, drag due to air friction, and the attractive forces of the magnets. For simplicity, we treat the magnets as magnetic point charges and assume that the length of the pendulum rod is much greater than the distance between the magnets, allowing us to describe the dynamics using a small-angle approximation. 

The resulting dimensionless equations of motion for the pendulum bob are
\begin{align} 
    \ddot{x} &= -\omega_0^2 x- a \dot{x}+\sum_{i=1}^3 \frac{\xt_i-x}{D\left(\xt_i, \yt_i\right)^3} \label{eq:x-dyn}, \\
    \ddot{y} &= -\omega_0^2 y- a \dot{y}+\sum_{i=1}^3 \frac{\yt_i-y}{D\left(\xt_i, \yt_i\right)^3} \label{eq:y-dyn},
\end{align}
where $(\xt_i,\yt_i)$ are the coordinates of the $i$th magnet, $\omega_0$ is the pendulum's natural frequency, and $a$ is the damping coefficient. Here, $D\left(\xt, \yt \right)$ denotes the distance between the bob and a given point $(\xt, \yt)$ in the magnets' plane:
\begin{equation}
D\left(\xt, \yt \right) = \sqrt{\left(\xt-x\right)^2+\left(\yt-y\right)^2+h^2},
\label{eq:distance}
\end{equation}
where $h$ is the bob's height above the plane. The system's four-dimensional state is thus $\x = \left(x, y, \dot{x}, \dot{y}\right)^T$.

We take the $(x, y)$ coordinates of the magnets to be 
$\left(\nicefrac{1}{\sqrt{3}}, 0\right)$,\,
$\left(\nicefrac{-1}{2 \sqrt{3}}, \nicefrac{-1}{2}\right)$,\,
and $\left(\nicefrac{-1}{2\sqrt{3}}, \nicefrac{1}{2}\right)$.
Unless stated otherwise, we set $\omega_0 = 0.5$, $a = 0.2$, and $h = 0.2$ in our simulations. These values are representative of all cases for which the magnetic pendulum has exactly three stable fixed points, corresponding to the bob being at rest and pointed toward one of the three magnets.

Previous studies have largely focused on chaotic dynamics as a stress test of RC's capabilities \cite{jaeger2004harnessing,pathak2017using,pathak2018model,carroll2018using,rafayelyan2020large,fan2020long,kim2021teaching,gauthier2021next,patel2023using}. 
Here we take a different approach. With non-zero damping, the magnetic pendulum dynamics is autonomous and dissipative, meaning \emph{all} trajectories must eventually converge to a fixed point. Except on a set of initial conditions of measure zero, this will be one of the three stable fixed points identified earlier. Yet predicting which attractor a given initial condition will go to can be far from straightforward, with the pendulum wandering in an erratic transient before eventually settling to one of the three magnets \cite{motter2013doubly}. This manifests as complicated basins of attraction with a ``pseudo'' (fat) fractal structure (\cref{fig:schematic}).
We can control the ``fractalness'' of the basins by, for example, varying the height of the pendulum $h$. This generates basins with tunable complexity to test the performance of (NG)RC.

\section{Implementation of Standard RC}
\label{sec:standard-rc-implementation}

Consider a dynamical system whose $n$-dimensional state $\mathbf{x}$ obeys a set of $n$ autonomous differential equations of the form
\begin{equation}
\dot{\x} = \f(\x).
\label{eq:continuous-dyn}
\end{equation}
In general, the goal of reservoir computing is to approximate the flow of Eq.~(\ref{eq:continuous-dyn}) in discrete time by a map of the form
\begin{equation}
\x_{t+1} = \F(\x_t).
\label{eq:discrete-dyn}
\end{equation}
Here, the index $t$ runs over a set of discrete times separated by $\Delta t$ time units of the real system,
where $\Delta t$ is a timescale hyperparameter generally chosen to be smaller than the characteristic timescale(s) of \cref{eq:continuous-dyn}. 

In standard RC, one views the state of the real system as a linear readout from an auxiliary \emph{reservoir system}, whose state is an $N_r$-dimensional vector $\res_t$. Specifically: 
\begin{equation}
    \x_t = \W \cdot \res_t,
\label{eq:standard-rc-readout}
\end{equation}
where $\W$ is an $n \times N_r$ matrix of trainable output weights. The reservoir system is generally much higher-dimensional ($N_r \gg n$), and its dynamics obey
\begin{equation}
    \res_{t+1} = \left(1 - \alpha\right) \res_t + \alpha f\left(\Wr \cdot \res_t + \Win \cdot \mathbf{u}_t + \bias \right).
\label{eq:rc-dyn}
\end{equation}
Here $\*W_r$ is the $N_r \times N_r$ reservoir matrix, $\Win$ is the $N_r \times n$ input matrix, and $\bias$ is an $N_r$-dimensional bias vector. The input $\*u_t$ is an $n$-dimensional vector that represents either a state of the real system ($\*u_t = \*x_t$) during training or the model's own output (${\*u_t = \*W \cdot \*r_t}$) during prediction. The nonlinear activation function $f$ is applied elementwise, where we adopt the standard choice of $f(\cdot) = \tanh(\cdot)$. 
Finally, $0 < \alpha \le 1$ is the so-called \emph{leaky coefficient}, which controls the inertia of the reservoir dynamics. 

In general, only the output matrix $\W$ is trained, with $\Wr$, $\Win$, and $\bias$ generated randomly from appropriate ensembles. We follow best practices \cite{lukovsevivcius2012practical} and previous studies in generating these latter components, specifically:
\begin{itemize}
    \item $\Wr$ is the weighted adjacency matrix of a directed Erd\H{o}s-R\'{e}nyi graph on $N_r$ nodes. The link probability is $0 < q \le 1$, and we allow for self-loops. We first draw the link weights uniformly and independently from $[-1, 1]$, and then normalize them so that $\Wr$ has a specified spectral radius $\rho > 0$. Here, $q$ and $\rho$ are hyperparameters.
    
    \item $\Win$ is a dense matrix, whose entries are initially drawn uniformly and independently from $[-1, 1]$. In the magnetic pendulum, the state $\x_t$ (and hence the input term $\mathbf{u}_t$ in \cref{eq:rc-dyn}) is of the form
    $\left(x, y, \dot{x}, \dot{y}\right)^T$. To allow for different characteristic scales of the position vs.\ velocity dynamics, we scale the first two columns of $\Win$ by $s_x$ and the last two columns by $s_v$, where $s_x, s_v > 0$ are scale hyperparameters.

    \item $\mathbf{b}$ has its entries drawn uniformly and independently from $[-s_b, s_b]$, where $s_b > 0$ is a scale hyperparameter.
\end{itemize}

\noindent \textbf{Training.} To train an RC model from a given initial condition $\x_0$, we first integrate the real dynamics (\ref{eq:continuous-dyn}) to obtain $\Ntrain$ additional states $\lbrace \x_t \rbrace_{t=1,\ldots,\Ntrain}$. We then iterate the reservoir dynamics (\ref{eq:rc-dyn}) for $\Ntrain$ times from ${\*r_0 = \*0}$, using the training data as inputs ($\mathbf{u}_t = \x_t$). This produces a corresponding sequence of reservoir states, $\lbrace \res_t \rbrace_{t=1,\ldots,\Ntrain}$. Finally, we solve for the output weights $\W$ that render \cref{eq:standard-rc-readout} the best fit to the training data using Ridge Regression with Tikhonov regularization:        
\begin{equation}
\*W = \*X \*R^T \left(\*R \*R^T + \lambda \mathbb{I}\right)^{-1}.
\label{eq:ridge-regression-rc}
\end{equation}
Here, $\*X$ ($\*R$) is a matrix whose columns are the $\x_t$ ($\res_t$) for $t=1,\ldots,\Ntrain$, $\mathbb{I}$ is the identity matrix, and $\lambda > 0$ is a regularization coefficient that prevents ill-conditioning of the weights, which can be symptomatic of overfitting the data. \\

\noindent \textbf{Prediction.} To simulate a trained RC model from a given initial condition $\x_0$, we first integrate the true dynamics (\ref{eq:continuous-dyn}) forward in time to obtain a total of $N_\text{warmup} \ge 0$ states $\lbrace \x_t \rbrace_{t=1,\ldots,N_\text{warmup}}$. During the first $N_\text{warmup}$ iterations of the discrete dynamics (\ref{eq:rc-dyn}), the input term comes from the real trajectory, i.e., $\mathbf{u}_t = \x_t$. Thereafter, we replace the input with the model's own output at the previous iteration ($\*u_t = \*W \cdot \*r_t$). This creates a closed-loop system from \cref{eq:rc-dyn}, 
which we iterate without further input from the real system.

\section{Critical Dependence of Standard RC on Warmup Time}
\label{sec:standard-rc-results}

\begin{figure}
    \centering
    \includegraphics{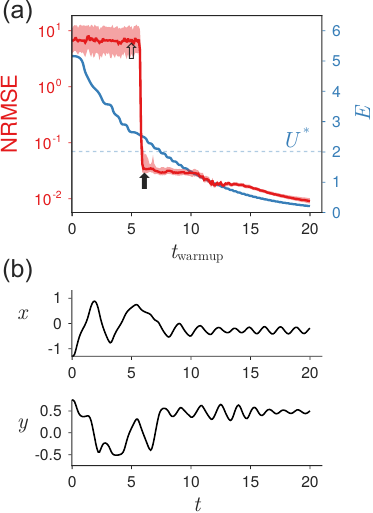}
    \caption{\textbf{Forecastability transition of standard RC.} The initial condition used for both training and prediction was $(x_0, y_0) = (-1.3, 0.75)$. The optimized RC hyperparameters for this initial condition are listed in \cref{table:rc-param-values}. We train an ensemble of 100 different RC models with these hyperparameters, then simulate each from the training initial condition using varying numbers of warmup points. In (a), the red line/bands denote the median/interquartile range of the resulting prediction NRMSE as a function of warmup time. The blue curves show the total mechanical energy $E$ of the training trajectory at the same times. We see a sharp drop in model error at $t_\text{warmup} \approx 6$, only shortly before $E$ falls below the height of the potential barriers ($U^*$, dashed line) separating the three wells of the magnetic pendulum. In (b), we overlay the $x$ and $y$ dynamics of the real system for comparison. This confirms that the ``critical'' warmup time in (a) aligns closely with the end of the transient. The arrows in (a) denote the warmup times used for the example simulations in \cref{fig:err_vs_warmup_example1}. }
    \label{fig:err_vs_warmup1}
\end{figure}

Though standard RC is extremely powerful, it is known to demand large warmup periods ($N_\text{warmup}$) in certain problems in order to be stable \cite{rohm2021model}. In principle, this could create a dilemma for the problem of basin prediction, as long warmup trajectories from the real system will generally be unavailable for initial conditions unseen during training. And even if such data were available, the problem could be rendered moot if the required warmup exceeds the transient period of the given initial condition \cite{gauthier2022learning}.
Here, we systematically test RC's sensitivity to the warmup time using the magnetic pendulum system. 

Our aim is to test standard RC under the most favorable conditions. Accordingly, we will train each RC model on a \emph{single} initial condition $\x_0 = (x_0, y_0, 0, 0)^T$ of the magnetic pendulum, and ask it to reproduce only the trajectory from that initial condition. Likewise, before training, we systematically optimize the RC hyperparameters for that initial condition via Bayesian optimization, seeking to minimize an objective function that combines both training and validation error. For details of this process, we refer the reader to \cref{sec:bayes_opt}. 

\begin{figure}
    \centering
    \includegraphics{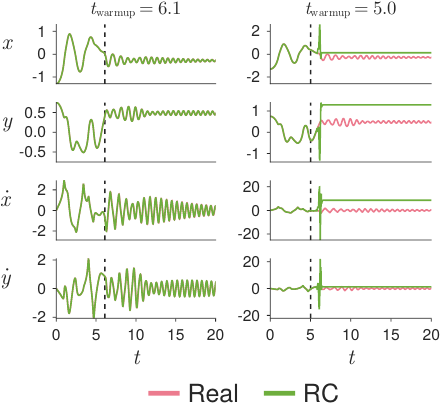}
    \caption{\textbf{Sensitivity of standard RC performance to warmup time.}
    We show example simulations from one RC realization in \cref{fig:err_vs_warmup1} with two different warmup times (dashed lines). The initial condition and optimized RC hyperparameters are the same as in \cref{fig:err_vs_warmup1}. 
    }
    \label{fig:err_vs_warmup_example1}
\end{figure}

In initial tests of our optimization procedure, we found it largely insensitive to the reservoir connectivity ($q$), with equally good training/validation performances achievable across a range of $q$ from $0.01$ to $1$. We likewise found little impact of the regularization coefficient over several orders of magnitude, with the optimizer frequently pinning $\lambda$ at the supplied lower bound of $10^{-8}$. Thus, in the interest of more fully exploring the most important hyperparameters, we fix $q = 0.03$ and $\lambda = 10^{-8}$. We then optimize the remaining five continuous hyperparameters ($\rho$, $s_x$, $s_v$, $s_b$, $\alpha$) over the ranges specified in \cref{table:optimizable-rc-params}. 

Throughout this section, we set $\Delta t = 0.02$, which is smaller than the characteristic timescales of the magnetic pendulum. We train each RC model on $\Ntrain = 4000$ data points of the real system starting from the given initial condition, which when paired with the chosen $\Delta t$ encompass both the transient dynamics and convergence to one of the attractors. We fix the reservoir size at $N_r = 300$, and we show that larger reservoir sizes do not alter our results in Supplemental Material.

Figure \ref{fig:err_vs_warmup1} shows the performance of an ensemble of RC realizations with optimized hyperparameters for the initial condition $(x_0, y_0) = (-1.2, 0.75)$. Specifically, we show the normalized root-mean-square error (NRMSE, see \cref{sec:nrmse}) between the real and RC-predicted trajectory as a function of warmup time ($t_\text{warmup} = N_\text{warmup} \cdot \Delta t$). In \cref{fig:err_vs_warmup1}(a) we observe a sharp transition around $t_\text{warmup}=6$. Before this point, we consistently have NRMSE $= \mathcal{O}(1)$, meaning the RC error is comparable to the scale of the real trajectory. But after the transition, the error is always quite small (NRMSE $\ll 1$). 

We can gain physical insight about this ``forecastability transition'' by analyzing the total mechanical energy of the training trajectory:
\begin{equation}
E = \frac{1}{2} \left(\dot{x}^2 + \dot{y}^2\right) + U(x, y).
\label{eq:energy}
\end{equation}
Here $U(x, y)$ is the potential corresponding to Eqs.~\ref{eq:x-dyn}-\ref{eq:y-dyn}, where we set $U = 0$ at the minima corresponding to the three attractors. Strikingly, the critical warmup time occurs only shortly before the energy drops below a critical value $U^*$---defined as the height of the potential barriers between the three wells [\cref{fig:err_vs_warmup1}(a)]. By this time, the system is unambiguously ``trapped'' near a specific magnet, making only damped oscillations thereafter [\cref{fig:err_vs_warmup1}(b)]. This suggests that even highly optimized RC models will fail to reproduce convergence to the correct attractor unless they have already been guided there by data from the real system.

We illustrate this further in \cref{fig:err_vs_warmup_example1}, showing example predictions from one RC realization considered above under two different warmup times: one above the critical value in \cref{fig:err_vs_warmup1}, and one below. Indeed, with sufficient warmup (left), the RC trajectory is a near-perfect match to the real one, both before and after the warmup period. But if the warmup time is even slightly less than the critical value (right), the model quickly diverges once the autonomous prediction begins. In this case, the model fails to reproduce convergence to \emph{any} fixed-point attractor, let alone the correct one, instead oscillating wildly. 

This pattern holds when we repeat our experiment for other initial conditions, re-optimizing hyperparameters and re-training an ensemble of RC models for each (Figs.~\ref{fig:err_vs_warmup2}--\ref{fig:err_vs_warmup_example3}). In all cases, we see the same sharp drop in RC prediction error at a particular warmup time (Figs.~\ref{fig:err_vs_warmup2},\ref{fig:err_vs_warmup3}). Without at least this much warmup time, the models fail to capture the real dynamics even qualitatively, often converging to an unphysical state with non-zero final velocity (Figs.~\ref{fig:err_vs_warmup_example2},\ref{fig:err_vs_warmup_example3}). Though there exist initial conditions that require shorter warmups---such as $(x_0, y_0) = (1.0, -0.5)$---this is only because those initial conditions have shorter transients. Indeed, there are other initial conditions---such as $(x_0, y_0) = (1.75, 1.6)$---that have longer transients and demand commensurately larger warmup times (Figs.~\ref{fig:err_vs_warmup3}, \ref{fig:err_vs_warmup_example3}). In no case have we observed the RC dynamics staying faithful to the real system unless the warmup is comparable to the transient period.

Note that the breakdown of RC with insufficient warmup time cannot be attributed to an insufficiently complex model vis-\`{a}-vis the only hyperparameter we have not optimized: the reservoir size ($N_r$). Indeed, we have repeated our experiment with reservoirs twice as large ($N_r = 600$). Even with optimized values of the other hyperparameters, we still see a sharp transition in the NRMSE at a warmup time comparable to the transient time (\cref{fig:err_vs_warmup_example_big}).

In sum, we have shown that standard RC is unsuitable for basin prediction in this representative multistable system. Specifically, RC models can only reliably reproduce convergence to the correct attractor when they have been guided to its vicinity. This is true even with the benefit of highly tuned hyperparameters (\cref{sec:bayes_opt}), and validation on only initial conditions seen during training.

For the remainder of the paper, we instead turn to Next Generation Reservoir Computing (NGRC). Though it is known that every NGRC model implicitly defines the connectivity matrix and other parameters of a standard RC model \cite{bollt2021explaining,gauthier2021next}, there is no guarantee that the two architectures would perform similarly in practice. In particular, NGRC is known to demand substantially less warmup time \cite{gauthier2021next}, potentially avoiding the ``catch-22'' identified here for standard RC. Can this cutting-edge framework succeed in learning the magnetic pendulum and other paradigmatic multistable systems?

\section{Implementation of NGRC}
\label{sec:ngrc-implementation}

We implement the NGRC framework following Refs.~\cite{gauthier2021next, gauthier2022learning}. In NGRC, the update rule for the discrete dynamics is taken as:
\begin{equation}
\x_{t+1} = \x_t + \W \cdot \g_t,
\label{eq:ngrc-dyn}
\end{equation}
where $\g_t$ is an $m$-dimensional \emph{feature vector}, calculated from the current state and $k - 1$ past states, namely
\begin{equation}
\g_t = \g\left(\x_t, \x_{t-1}, \ldots , \x_{t-k+1} \right).
\label{eq:features}
\end{equation}

\noindent Here, $k \ge 1$ is a hyperparameter that governs the amount of memory in the NGRC model, and $\W$ is an $n \times m$ matrix of trainable weights.

We elaborate on the functional form of the feature embedding $\g$ below. But in general, the features can be divided into three groups: (i) one constant (bias) feature; (ii) $\mlin = nk$ linear features, corresponding to the components of $\lbrace \x_t, \x_{t-1}, \ldots, \x_{t-k+1} \rbrace$; and finally (iii) $\mnonlin$ nonlinear features, each a nonlinear transformation of the linear features. The total number of features is thus $m = 1 + \mlin + \mnonlin$.  \\

\noindent \textbf{Training}. Per \cref{eq:ngrc-dyn}, training an NGRC model amounts to finding values for the weights $\W$ that give the best fit for the discrete update rule
\begin{equation}
\y_t = \W \cdot \g_t,
\label{eq:train-target}
\end{equation}
where $\y_t = \x_{t+1} - \x_t$. Accordingly, we calculate pairs of inputs ($\g_t$) and next-step targets ($\y_t$) over $\Ntraj \ge 1$ training trajectories from the real system (\ref{eq:continuous-dyn}), each of length $\Ntrain + k$.
We then solve for the values of $\W$ that best fit \cref{eq:train-target} in the least-squares sense via regularized Ridge regression, namely
\begin{equation}
\W = \Y \G^T \left(\G \G^T + \lambda \mathbb{I}\right)^{-1}.
\label{eq:ridge-regression}
\end{equation}
Here $\Y$ ($\G$) is a matrix whose columns are the $\y_t$ ($\g_t$). The regularization coefficient $\lambda$ plays the same role as in standard RC [cf.~\cref{eq:ridge-regression-rc}]. \\

\noindent \textbf{Prediction}. To simulate a trained NGRC model from a given initial condition $\x_0$, we first integrate the true dynamics (\ref{eq:continuous-dyn}) forward in time to obtain the additional $k - 1$ states needed to perform the first discrete update according to Eqs.~(\ref{eq:ngrc-dyn})-(\ref{eq:features}). 
This is the warmup period for the NGRC model.
Thereafter, we iterate Eqs.~(\ref{eq:ngrc-dyn})-(\ref{eq:features}) as an autonomous dynamical system, with each output becoming part of the model's input at the next time step. Thus in contrast to training, the model receives no data from the real system during prediction except the $k - 1$ ``warm-up'' states. \\

There is a clear parallel between NGRC \cite{bollt2021explaining,gauthier2021next} and statistical forecasting methods \cite{billings2013nonlinear} such as nonlinear vector-autoregression (NVAR).
However, as noted in Ref.~\cite{jaurigue2022connecting}, the feature vectors of a typical NGRC model usually have far more terms than NVAR methods, as the latter was designed with interpretability in mind. 
It is the use of a library of many candidate features---in addition to other details like the typical training methods employed---that sets NGRC apart from classic statistical forecasting approaches. In this way, NGRC also resembles the Sparse Identification of Nonlinear Dynamics (SINDy) framework \cite{brunton2016discovering}.
The differences here are the intended tasks (finding parsimonious models vs.\ fitting the dynamics), the optimization schemes (LASSO vs.\ Ridge regression), and NGRC's inclusion of delayed states (generally no delayed states for SINDy).

\begin{table*}
\small
\begin{tabular}{ccccc}
\toprule
Model & Nonlinear Features & Example Term(s) & \makecell{Addl. Hyperparameters} & \makecell{$m_\textrm{nonlin}$} \\
\midrule
I & Polynomials & \makecell{$x_t^2 \dot{y}_t$ \\ $y_{t-2} \dot{y}_{t-1} \dot{x}_t$} & max.~degree $d_{\max}$ & $\displaystyle \nsum[1.6]_{d=2}^{d_{\max}} \multiset{4 k}{d}$ \\
\midrule
II & \makecell{Radial Basis \\ Functions} & 
$\displaystyle \frac{1}{\left(\lVert \mathbf{r}_t - \mathbf{c}_i \rVert^2 + h^2\right)^{3/2}}$ & centers $\lbrace \mathbf{c}_i \rbrace_{i=1, \ldots, \NRBF}$ & $\NRBF k$ \\
\midrule
III & \makecell{Pendulum Forces} & \makecell{$\displaystyle \frac{\yt_i - y_t}{D\left(\xt_i, \yt_i\right)^3}$ \\ 
}& & $6k$ \\
\bottomrule
\end{tabular}

\caption{\textbf{Summary of NGRC models constructed for the magnetic pendulum system}. For each model described in \cref{sec:nonlinearity_pendulum}, we provide examples of the nonlinear features, their total number ($\mnonlin$), and any additional hyperparameters.
(I) Here, $\multiset{a}{b}$ denotes the number of ways to choose $b$ items (with replacement) from a set of size $a$. 
(II) Here, $\bm{r}_t = \left(x_t, y_t\right)$ are the position coordinates at time $t$, and $\bm{c}_i$ is the $i$th RBF center in 2D, whose $x$ and $y$ coordinates are drawn independently from a uniform distribution over $[-1.5, 1.5]$.
(III) Here, $(\xt_i, \yt_i)$ are coordinates of the $i$th magnet in the real system ($i = 1,2,3$), and $D(\xt, \yt)$ is as in \cref{eq:distance}.
}
\label{table:models}
\end{table*}

\section{Sensitive dependence of NGRC performance on readout nonlinearity}
\label{sec:nonlinearity_pendulum}

The importance of careful feature selection is well-appreciated for many machine learning 
frameworks \cite{rahimi2007random,brunton2016discovering}.
Yet one major appeal of NGRC is that the choice of nonlinearity is considered to be of secondary importance; in many systems studied to date, one can often bypass the feature selection process by adopting some generic nonlinearities (e.g., low-order polynomials).
Indeed, applications of NGRC to chaotic benchmark systems have shown good results even when the features do not include all nonlinearities in the underlying ODEs \cite{gauthier2021next,shahi2022prediction}. 
But can we expect this to be true in general?

\begin{figure}[tb]
\centering
\includegraphics[width=.99\columnwidth]{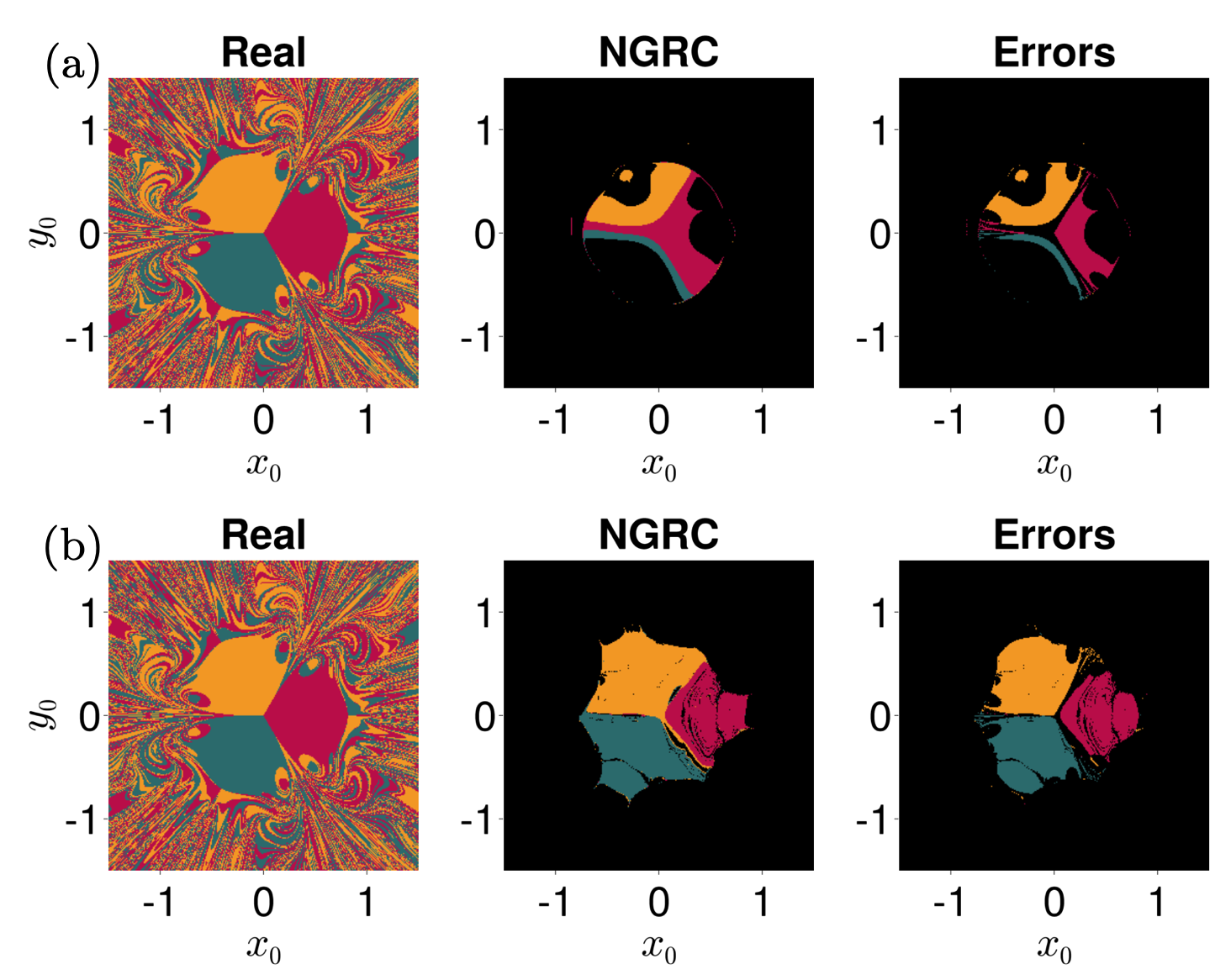}
\caption{
\textbf{NGRC models with polynomials as their nonlinearity fail to capture the basins of the magnetic pendulum system.} 
We tested the basin predictions made by NGRC models with the number of time-delayed states up to $k=5$ and the maximum degree of the polynomial up to $d_{\max}=5$. 
Two representative predictions are shown for (a) $k=5$, $d_{\max}=3$ and (b) $k=3$, $d_{\max}=5$. 
The left panels show the ground-truth basins of the magnetic pendulum system;
The middle panels show the basins identified by the NGRC models, where black points denote initial conditions from which the NGRC trajectories diverge to infinity;
The right panels show the correctly identified basins in colors and the misidentified basins in black.
The hyperparameters used in this case are $\Delta t = 0.01$, $\lambda= 1$, $\Ntraj = 100$, and $\Ntrain = 5000$.
}
\label{fig:polynomial}
\end{figure}

\begin{figure*}[tbh!]
\centering
\includegraphics[width=1\linewidth]{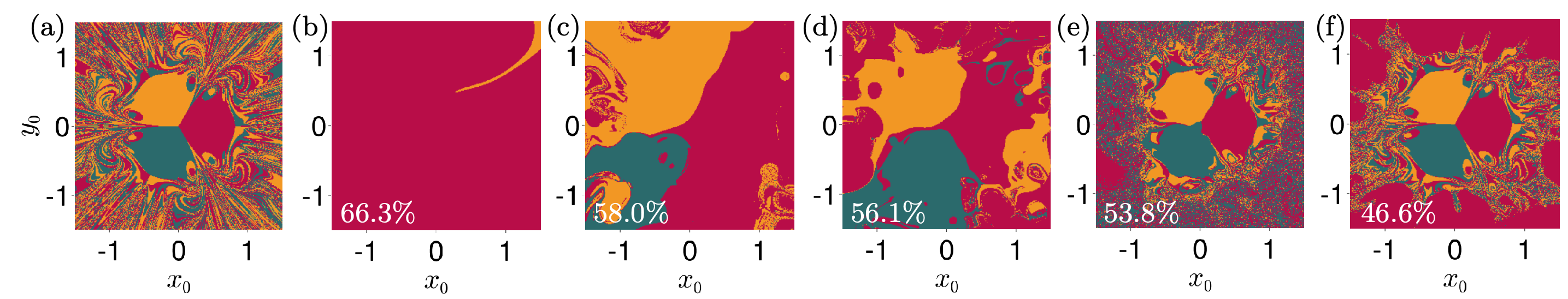}
\caption{
\textbf{NGRC models with radial basis functions as their readout nonlinearity struggle to capture the basins of the magnetic pendulum system.}
We tested the basin predictions made by NGRC models whose nonlinear features include $\NRBF$ radial basis functions. 
Panel (a) shows the ground truth, and the rest of the panels show representative NGRC predictions for (b) $\NRBF = 10$, (c) $\NRBF = 50$, (d) $\NRBF = 100$, (e) $\NRBF = 500$, and (f) $\NRBF = 1000$.
The error rates of the predictions are indicated in the lower left corners.
The solutions no longer blow up as they did for the polynomial nonlinearities in Model I, but the NGRC models still struggle to capture the basins even qualitatively.
Even at $\NRBF = 1000$, only the most prominent features of the basins around the origin are correctly identified.
The other hyperparameters used are $\Delta t = 0.01$, $\lambda = 1$, $k=2$, $\Ntraj = 100$, and $\Ntrain = 5000$.
}
\label{fig:radial}
\end{figure*}

Here, we test NGRC's sensitivity to the choice of feature embedding $\g$ (i.e., readout nonlinearity) in the basin prediction problem. Specifically, we compare the performance of three candidate NGRC models, in which the nonlinearities are:
\begin{enumerate}[label=\Roman*.]
    \item \emph{Polynomials}, specifically all unique monomials formed by the $4k$ components of $\lbrace \x_t, \x_{t-1}, \ldots , \x_{t-k+1} \rbrace$, with degree between 2 and $d_{\max}$.
    
    \item As set of $\NRBF$ \emph{Radial Basis Functions} (RBF) applied to the position coordinates $\bm{r} = \left(x, y\right)$ of each of the $k$ states.  The RBFs have randomly-chosen centers and a kernel function with shape and scale similar to the magnetic force term.
    
    \item The exact nonlinearities in the magnetic pendulum system. Namely, the $x$ and $y$ components of the magnetic force for each magnet, evaluated at each of the $k$ states. 
\end{enumerate}

The details of each model are summarized in \cref{table:models}. Recall that in addition to their unique nonlinear features, all models contain one constant feature (set to 1 without loss of generality) and $4k$ linear features.

Models I-III represent a hierarchy of increasing knowledge about the real system. In Model I, we assume complete ignorance, hoping that the real dynamics are well-approximated by a truncated Taylor series. In Model II, we acknowledge that this is a Newtonian central force problem and even the shape/scale of that force, but plead ignorance about the locations of the point sources. Finally, in Model III, we assume perfect knowledge of the system that generated the time series. 
Between the linear and nonlinear features, Model III includes all terms in Eqs.~\ref{eq:x-dyn}-\ref{eq:y-dyn}. 

Our principal question is: How well each NGRC model can reproduce the basins of attraction of the magnetic pendulum and in turn predict its long-term behavior? We focus on the 2D region of initial conditions depicted in \cref{fig:schematic}, in which the pendulum bob is released from rest at position $(x_0, y_0)$, with $-1.5 \le x_0, y_0 \le 1.5$. We train each model on $\Ntraj$ trajectories generated by Eqs.~\ref{eq:x-dyn}-\ref{eq:y-dyn} from initial conditions sampled uniformly and independently from the same region. We then compare the basins predicted by each trained NGRC model with those of the real system (\cref{sec:basin_prediction}). We define the \emph{error rate} ($p$) as the fraction of initial conditions for which the basin predictions disagree. \\

\begin{figure*}[t]
\centering
\includegraphics{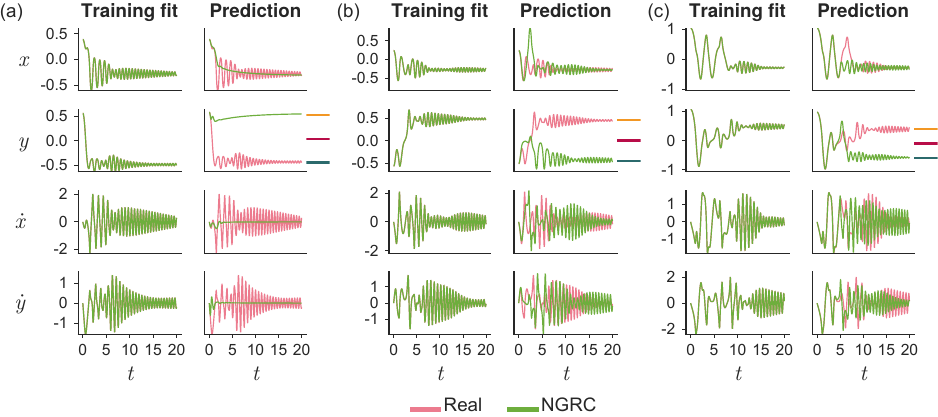}
\caption{
\textbf{NGRC models frequently mis-forecast even the initial conditions they were trained on.} The panels correspond to (a) Model I, with $d_{\max}=3$; (b) Model II, with $\NRBF = 500$; and (c) Model III, with no uncertainty. Each model was trained on trajectories from $\Ntraj = 100$ initial conditions. For each model, we show one such initial condition for which NGRC (green) predicts the wrong basin, despite an excellent fit to the corresponding ground-truth training trajectory (pink). The left column of each panel shows the training fit.
The right column shows the autonomous NGRC simulation from the same initial condition. 
The three magnets can be distinguished by their $y$ coordinates [cf.~\cref{fig:schematic}], allowing us to indicate which one a given trajectory approaches via the three colored lines beside the second row in each panel.
In each case, the training fit is impeccable, with the two curves overlapping to within visual resolution (left).
Yet when the NGRC model is run \emph{autonomously} from the same initial condition, it quickly diverges from the ground truth, eventually going to an incorrect attractor (right).
For all models, we set the other hyperparameters as $\Delta t = 0.01$, $\lambda = 1$, $k=2$, and $\Ntrain = 5000$.
}
\label{fig:train-vs-predict-fail}
\end{figure*}

\noindent \textbf{Model I (Polynomial Features)}. For NGRC models equipped with polynomial features, excellent training fits can be achieved (Figs.~\ref{fig:train-vs-predict-fail}, \ref{fig:training_fit} and \ref{fig:train-vs-predict-success}). Despite this, the models struggle to reproduce the qualitative dynamics of the magnetic pendulum, let alone the basins of attraction. 

\Cref{fig:polynomial}(a) shows representative NGRC basin predictions made by Model I using $k=5$, $d_{\max}=3$. 
For the vast majority of initial conditions, the NGRC trajectory does not converge to any of the three attractors, instead diverging to (numerical) infinity in finite time (black points in the middle panels of \cref{fig:polynomial}). 
Modest improvements can be obtained by including polynomials up to degree $d_{\max} =5$ (with $k = 3$) as shown in \cref{fig:polynomial}(b). 
But even here, the model succeeds only at learning the part of each basin in the immediate vicinity of each attractor. 

Unfortunately, eking out further improvements by increasing the complexity of the NGRC model becomes computationally prohibitive.  
When $k = 3$ and $d_{\max} = 5$, for example, the model already has $m = 6,188$ features. Likewise, the feature matrix $\G$ used in training
has hundreds of millions of entries. 
With higher values of $k$ and/or $d_{\max}$, the model becomes too expensive to train and simulate on a standard computer. 

To ensure the instability of the polynomial NGRC models is not caused by a poor choice of hyperparameters, we have repeated our experiments for a wide range of time resolutions $\Delta t$, training trajectory lengths $\Ntrain$, numbers of training trajectories $\Ntraj$ (\cref{fig:error_n_trj}), and values for regularization coefficient $\lambda$ spanning ten orders of magnitude (\cref{fig:error_lambda}). 
The performance of Model I was not significantly improved in any case. \\

\noindent \textbf{Model II (Radial Basis Features)}. For NGRC models using radial basis functions as the readout nonlinearity, the solutions no longer blow up as they did in Model I above. This is encouraging though perhaps unsurprising, as the RBFs are much closer to the nonlinearity in the original equations describing the magnetic pendulum system. 
Unfortunately, the accuracy of the NGRC models in predicting basins remains poor.

\Cref{fig:radial} shows representative NGRC basin predictions as the number of radial basis functions is increased from $\NRBF = 10$ to $\NRBF = 1000$. In all cases, fits to the training data are impeccable, with the root-mean-square error (RMSE) ranging from $0.003$ ($\NRBF=10$) to $0.0005$ ($\NRBF = 1000$).
As more and more RBFs are included, the predictions can be visibly improved, but this improvement is very slow.
For example, at $\NRBF = 1000$ (\cref{fig:radial}f),
the trained model predicts the correct basin for only 53.4\% of the initial conditions under study ($p = 0.466$). Moreover, most of this accuracy is attributable to the large central portions of the basins near the attractors, in which the dynamics are closest to linear. Outside of these regions, the NGRC basin map may appear fractal, but the basin predictions themselves are scarcely better than random guesses. This deprives us of accurate forecasts in precisely the regions of the phase space where the outcome is most in question.

As with the polynomial case above, we have repeated our experiments for a wide range of hyperparameters to rule out overfitting or poor model calibration (\cref{fig:error_n_trj} and \cref{fig:error_lambda}). The accuracy of Model II cannot be meaningfully improved with any of these changes. \\

\begin{figure}[t]
\centering
\includegraphics[width=.99\columnwidth]{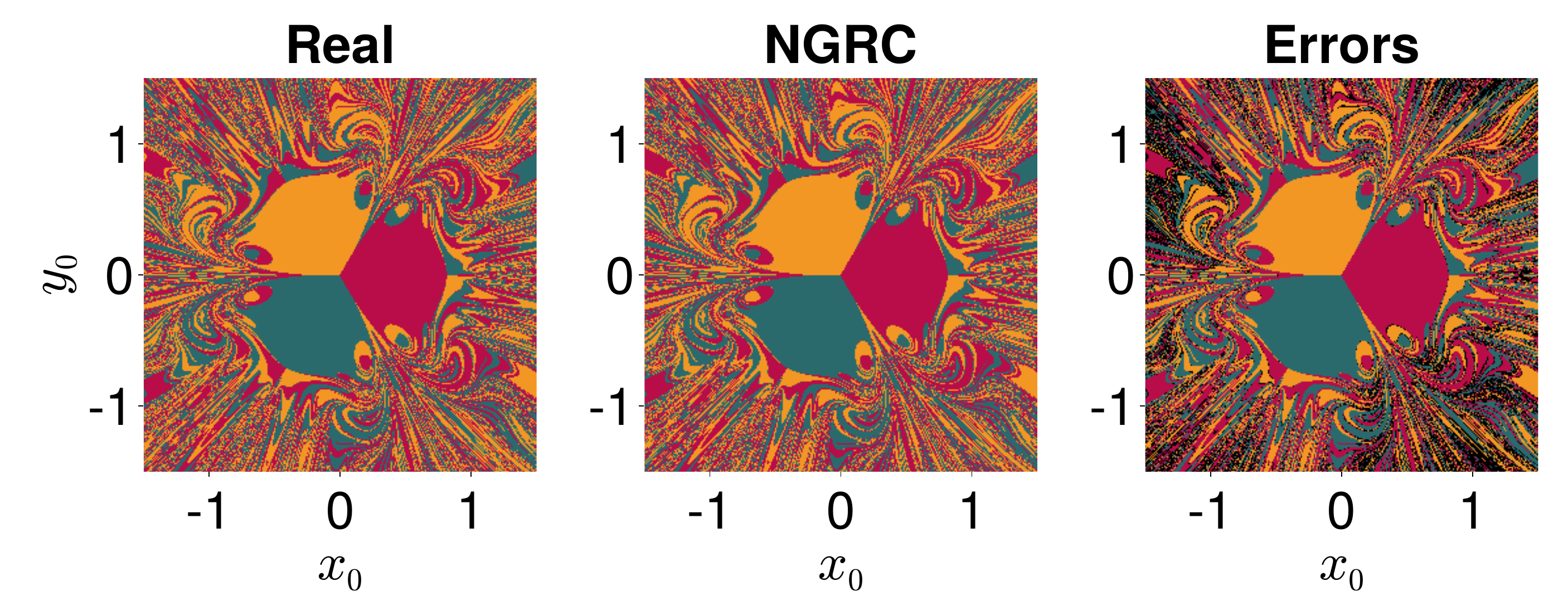}
\caption{
\textbf{NGRC models trained on a single trajectory can accurately capture all three basins when the exact nonlinearity from the magnetic pendulum system is adopted.}
The hyperparameters used are $\Delta t = 0.01$, $\lambda = 10^{-4}$, $k=4$, $\Ntraj = 1$, and $\Ntrain = 1000$, which achieves an error rate $p$ of $15\%$.
No systematic optimization was performed to find these parameters.
For example, by lowering $\Delta t$ to $0.0001$ and increasing $\Ntrain$ to $100000$, we can further improve the accuracy to over $98\%$.
}
\label{fig:single_trj}
\end{figure}

\noindent \textbf{Model III (Exact Nonlinearities)}. We next test NGRC models equipped with the exact form of the nonlinearity in the magnetic pendulum system, namely the force terms in Eqs.~(\ref{eq:x-dyn})-(\ref{eq:y-dyn}). This time, the NGRC models can perform exceptionally well. \Cref{fig:error_delta_t} shows the error rate of NGRC basin predictions as a function of the time resolution $\Delta t$. Without any fine-tuning of the other hyperparameters, NGRC models already achieve a near-perfect accuracy of $98.6\%$, provided $\Delta t$ is sufficiently small.

Astonishingly, Model III's predictions remain highly accurate even when it is trained on a \emph{single} trajectory ($\Ntraj = 1$) from a randomly-selected initial condition.
Here, NGRC can produce a map of all three basins that is very close to the ground truth ($85.0\%$ accuracy, \cref{fig:single_trj}), despite seeing data from only \emph{one} basin during training. This echoes previous results reported for the Li-Sprott system \cite{gauthier2022learning}, in which NGRC accurately reconstructed the basins of all three attractors (two chaotic, one quasiperiodic) from a single training trajectory. But how can we account for this night-and-day difference with the more system-agnostic models (I \& II), which showed poor performance despite 100-fold more training data?

The answer lies in the construction of the NGRC dynamics.
In possession of the exact terms in the underlying differential equations, \cref{eq:ngrc-dyn} can---by a suitable choice of the weights $\W$---emulate the action of a numerical integration method from the linear-multistep family \cite{butcher2016numerical}, whose order depends on $k$. 
When $k = 1$, for example, \cref{eq:ngrc-dyn} can mimic an Euler step. 
Thus, with a sufficiently small step size ($\Delta t$), it is not surprising that an NGRC model equipped with exact nonlinearities can accurately reproduce the dynamics of almost any differential equations.

This observation might explain the stellar performance of NGRC in forecasting specific chaotic dynamics like the Lorenz \cite{gauthier2021next} and Li-Sprott systems \cite{gauthier2022learning}. 
The nonlinearities in these systems are quadratic, meaning that so long as $d_{\max} \ge 2$, Model I can \emph{exactly} learn the underlying vector field. The only information to be learned is the coefficient ($\W$) that appears before each (non)linear term ($\g$) in the ODEs. This in turn could explain why a single training trajectory suffices to convey information about the phase space as a whole.\\

\begin{figure}[t]
\centering
\includegraphics[width=.99\columnwidth]{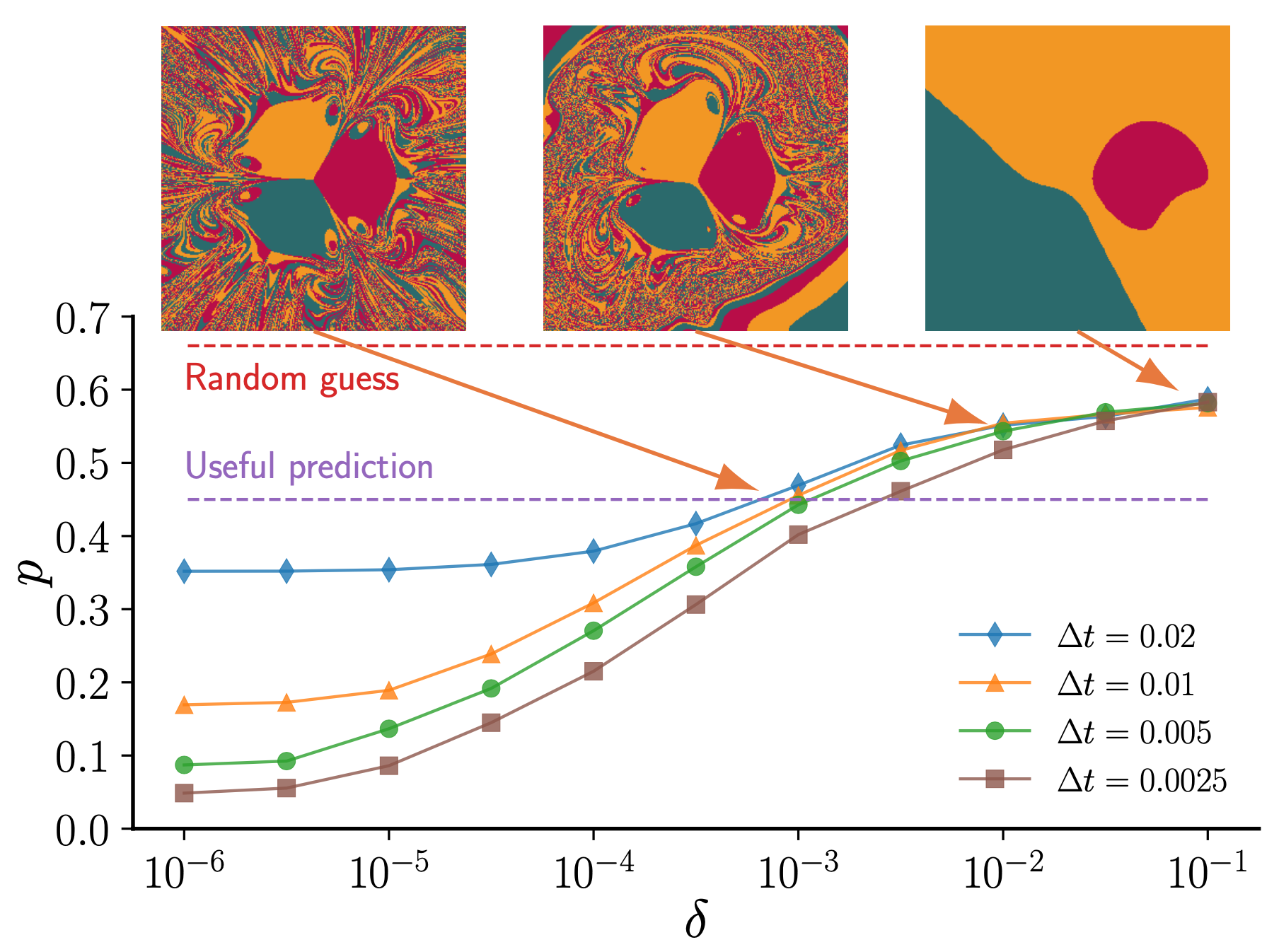}
\caption{
\textbf{NGRC basin prediction accuracy when using the exact nonlinearity from the pendulum equations but with small uncertainties.}
Here, the NGRC models adopt the exact nonlinearity in the magnetic pendulum system, except that the coordinates of the magnets are perturbed by amounts uniformly drawn from $[-\delta,\delta]$. 
Each data point is obtained by averaging the error rate $p$ over $10$ independent realizations.
We see that even a small uncertainty on the order of $\delta=10^{-5}$ can have a noticeable impact on the accuracy of basin predictions.
For $\delta>10^{-2}$, the NGRC predictions become unreliable, approaching the $66.6\%$ failure rate of random guesses.
Three representative NGRC-predicted basins are shown for $\delta=10^{-3}$, $\delta=10^{-2}$, and $\delta=10^{-1}$, respectively (all with $\Delta t = 0.01$).
We consider predictions with $p<0.45$ as useful since these in general produce basins that are visually similar to the ground truth.
The other hyperparameters used are $\lambda = 1$, $k=2$, $\Ntraj = 100$, and $\Ntrain = 5000$.
}
\label{fig:error_delta}
\end{figure}

\noindent \textbf{Model III with uncertainty}. Considering the wide gulf in performance between NGRC models equipped with exact nonlinearity and those equipped with polynomial/radial nonlinearity, it is natural to wonder whether there are some other smart choices of nonlinear features that perform well enough without knowing the exact nonlinearity. 

To explore this possibility, we consider a variant of Model III in which we introduce small uncertainties in the nonlinear features, perturbing the assumed coordinates of each magnet by small amounts drawn uniformly and independently between $\left [-\delta, \delta\right]$. Here $\delta $ is a hyperparameter much smaller than the characteristic spatial scale in this system ($\delta \ll 1$). We train the model on $\Ntraj = 100$ trajectories from the (unperturbed) real system, then measure how NGRC models perform in the presence of uncertainty about the exact nonlinearity.

In \cref{fig:error_delta}, we see that even a $\sim1\%$ mismatch ($\delta=0.01$) in the coordinates of the magnets $(\tilde{x}_i,\tilde{y}_i)$ is enough to make the accuracy of NGRC predictions plunge from almost $100\%$ to below $50\%$ (recall that even random guesses have an accuracy of $33.3\%$).
This extreme sensitivity of NGRC performance to perturbations in the readout nonlinearity suggests that any function other than the exact nonlinearity is unlikely to enable reliable basin predictions in the NGRC model. \\

\par \noindent \textbf{Training vs.\ prediction divergence}. In all models considered, we have seen that excellent fits to the training data do not guarantee accurate basin predictions for the rest of the phase space. But surprisingly, NGRC models can predict the wrong basin even for the precise initial conditions on which they were trained.

For each of Models I-III, \cref{fig:train-vs-predict-fail} shows one example training trajectory for which the model attains a near-perfect fit to the ground truth, but the NGRC trajectory from the same initial condition nonetheless goes to a different attractor. We can rationalize this discrepancy by considering the difference between the training and prediction phases as described in \Cref{sec:ngrc-implementation}. During training, NGRC is asked to calculate the next state given the $k$ most recent states from the ground truth data. In contrast, during prediction, the model must make this forecast based on its own (autonomous) trajectory. This permits even tiny errors to compound over time, potentially driving the dynamics to the wrong attractor. Though \cref{fig:train-vs-predict-fail} shows only one example for each model, these cases are quite common, regardless of the exact hyperparameters used \footnote{We did not observe any other attractors other than the three ground-truth fixed points and infinity for all NGRC models considered. The absence of more complicated attractors (compared to RC) is likely due to the simpler architecture of NGRC and the dissipativity of the real dynamics (which NGRC models can learn directly via the linear features).}. 

Moreover, in \cref{fig:train-vs-predict-success}, we show that even when the NGRC model predicts the correct attractor for a given training initial condition, the intervening transient dynamics can deviate significantly from the ground truth.
This is especially common and pronounced for NGRC models with polynomial or radial nonlinearities. 
In particular, the transient time---how long it takes to come close to the given attractor---can be much larger or smaller than in the real system.
As such, reaching the correct attractor does not necessarily imply that an NGRC model has learned the true dynamics from a given training initial condition. 
To say nothing of the (uncountably many) other initial conditions unseen during training. \\

\par \noindent \textbf{Influence of basin complexity.} As motivated earlier, the magnetic pendulum is a hard-to-predict system because of its complicated basins of attraction, regardless of the exact parameter values used. And indeed, we see the same sensitivity of NGRC performance to readout nonlinearity for other parameter values, such as $h=0.3$ and $h=0.4$ (\cref{fig:error_h}).

As the height of the pendulum $h$ is increased, the basins do tend to become less fractal-like.
In \cref{fig:h_dependence}, we vary the value of $h$ and show that NGRC models trained with polynomials fail even for the most regular basins ($h=0.4$).  On the other hand, NGRC models trained with radial basis functions see their performance improve significantly as the basins become simpler. As expected, NGRC models equipped with exact nonlinearity successfully capture the basins for all values of $h$ studied.

\section{Predicting high-dimensional basins with NGRC}
\label{sec:kuramoto}

How general are the results presented in \cref{sec:nonlinearity_pendulum}?
Could the magnetic pendulum be pathological in some unexpected way, with low-order polynomials or other generic features sufficing as the readout nonlinearity for most dynamical systems of interest?
To address this possibility, we investigate another paradigmatic multistable system---identical Kuramoto oscillators with nearest-neighbor coupling \cite{wiley2006size,delabays2017size,zhang2021basins}:
\begin{equation}
	\dot{\theta}_i = \sin(\theta_{i+1}-\theta_i) + \sin(\theta_{i-1}-\theta_i), \quad i = 1, \ldots, n,
	\label{eq:kuramoto_nn}
\end{equation}
where we assume a periodic boundary condition, so $\theta_{n+1} = \theta_1$ and $\theta_{0} = \theta_n$. 
Here $n$ is the number of oscillators and hence the dimension of the phase space, and $\theta_i(t) \in [0, 2\pi)$ is the phase of oscillator $i$ at time $t$. 

Aside from being well-studied as a model system: the Kuramoto system has two nice features. First, its sine nonlinearities are more ``tame'' than the algebraic fractions in the magnetic pendulum, helping to untangle whether the sensitive dependence observed in \Cref{sec:nonlinearity_pendulum} afflicts only specific nonlinearities. Second, we can easily change the dimension of \cref{eq:kuramoto_nn} by varying $n$, allowing us to test NGRC on high-dimensional basins. 

For $n>4$, \cref{eq:kuramoto_nn} has multiple attractors in the form of \emph{twisted states}---phase-locked configurations in which the oscillators' phases make $q$ full twists around the unit circle, satisfying $\theta_i = 2 \pi i q/n + C$.  Here $q$ is the winding number of the state \cite{wiley2006size}.  
Twisted states are fixed points of \cref{eq:kuramoto_nn} for all $q$, but only those with $|q|<n/4$ are stable \cite{delabays2017size}. The corresponding basins of attraction can be highly complicated \cite{zhang2021basins}, though not fractal-like as in the magnetic pendulum system. 

Similar to \cref{sec:nonlinearity_pendulum}, we consider three classes of readout nonlinearities assuming increasing knowledge of the underlying system:
\begin{enumerate}
    \item Monomials spanned by the $nk$ oscillator states in $\Theta = \lbrace \bm{\theta}_t, \bm{\theta}_{t-1}, \ldots , \bm{\theta}_{t-k+1} \rbrace$ , with degree between 2 and $d_{\max}$.
    \item Trigonometric functions of all scalars in $\Theta$, consisting of $\sin(\ell\theta_i)$ and $\cos(\ell\theta_i)$ for all $i$ and for integers  $1 \le \ell \le \ell_{\max}$.
    \item The exact nonlinearity in \cref{eq:kuramoto_nn}, namely ${\sin(\theta_i - \theta_{j})}$ for all pairs of connected nodes $i$ and $j$.
\end{enumerate}

\begin{figure}[t]
\centering
\includegraphics[width=.99\columnwidth]{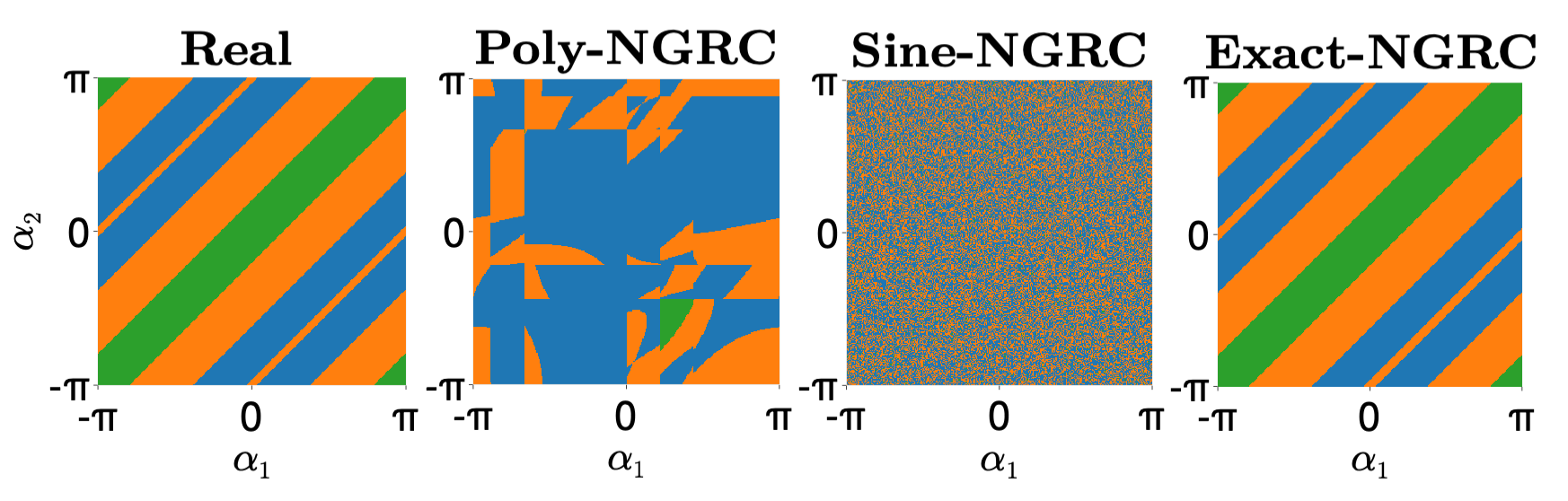}
\caption{
\textbf{Predicting basins of a Kuramoto oscillator network with NGRC.}
We show representative NGRC predictions for basins of $n=9$ locally-coupled Kuramoto oscillators.
Here, we select a 2D slice of the phase space centered at the twisted state with $q=2$.
Basins are color-coded by the absolute winding number $|q|$ of the corresponding attractor (blue: $|q|=0$; orange: $|q|=1$; green: $|q|=2$). 
Despite the simple geometry of the basins and extensive optimization of hyperparameters, NGRC models with polynomial nonlinearity ($d_{\max}=2$) or trigonometric nonlinearity ($\ell_{\max}=5$) have accuracies that are comparable to random guesses.
In contrast, with exact nonlinearity, NGRC predictions are consistently over $99\%$ correct.
The other hyperparameters are $\Delta t = 0.01$, $\lambda = {10}^{-5}$, $k=2$, $\Ntraj = 1000$, and $\Ntrain = 3000$.
}
\label{fig:kuramoto_n=9}
\end{figure}

To test the performance of different NGRC models on the Kuramoto system, we first set $n=9$ and use them to predict basins in a two-dimensional (2D) slice of the phase space. 
Specifically, we look at slices spanned by $\bm{\theta}_0 + \alpha_1\bm{P}_1 + \alpha_2\bm{P}_2$, $\alpha_i \in (-\pi,\pi]$.
Here, $\bm{P}_1$ and $\bm{P}_2$ are $n$-dimensional binary orientation vectors, while $\bm{\theta}_0$ is the base point at the center of the slice. 

\Cref{fig:kuramoto_n=9} shows results for orientation vectors given by
\begin{equation*}
    \bm{P}_1=[1,0,1,0,1,0,1,0,1], \quad \bm{P}_2=[0,1,0,1,0,1,0,1,0],
\end{equation*}
with $\bm{\theta}_0$ representing the $2$-twist state.
We can see that NGRC models with polynomial nonlinearity and trigonometric nonlinearity fail utterly at capturing the simple ground-truth basins.
This is despite an extensive search over the hyperparameters $\Delta t$, $\lambda$, $d_{\max}$, and $\ell_{\max}$.
On the other hand, the NGRC model with exact nonlinearity gives almost perfect predictions for a wide range of hyperparameters.
The hyperparameters in \cref{fig:kuramoto_n=9} are chosen so that trajectories predicted by the polynomial-NGRC model do not blow up.

\begin{figure}[t]
\centering
\includegraphics[width=\columnwidth]{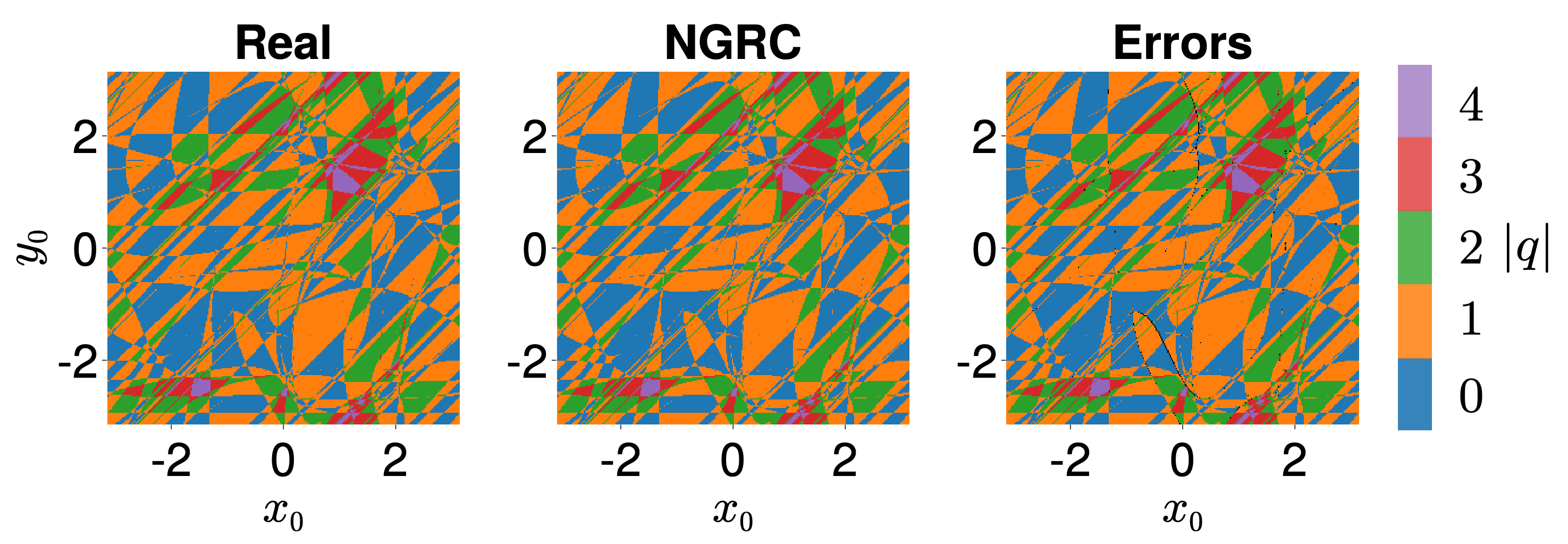}
\caption{
\textbf{NGRC with exact nonlinearity can accurately predict high-dimensional basins.}
Here we train an NGRC model (equipped with exact nonlinearity) to predict the high-dimensional basins of $n=83$ locally-coupled Kuramoto oscillators.
To test the NGRC performance, we randomly select a 2D slice of the $83$-dimensional phase space and compare the predicted basins with the ground truth.
Basins are color-coded by the absolute winding number $|q|$ of the corresponding attractor. 
Despite the fragmented and high-dimensional nature of the basins, NGRC captures the intricate basin geometry with ease.
Without deliberate optimization of the hyperparameters, NGRC can already achieve over $97\%$ accuracy.
The hyperparameters used are $\Delta t = 0.01$, $\lambda = {10}^{-5}$, $k=2$, $\Ntraj = 1000$, and $\Ntrain = 3000$.
}
\label{fig:kuramoto_n=83}
\end{figure}

Next, we show that the NGRC model with exact nonlinearity can predict basins in much higher dimensions and with more complicated geometries.
In \cref{fig:kuramoto_n=83}, we set $n=83$ and choose $\bm{\theta}_0$ to be a random point in the phase space.
The $n$-dimensional binary orientation vectors $\bm{P}_1$ and $\bm{P}_2$ are constructed by randomly selecting $\lfloor n/2 \rfloor$ components to be $1$ and the rest of the components are $0$.
(The results are not sensitive to the particular realizations of $\bm{P}_1$ and $\bm{P}_2$.)
Using the same hyperparameters as in \cref{fig:kuramoto_n=9}, the NGRC model achieves an accuracy of $97.5\%$.
Visually, one would be hard-pressed to find any difference between the predicted basins and the ground truth.

\section{Discussion}
\label{sec:end}

When can we claim that a machine learning model like RC has ``learned'' a dynamical system?
One basic requirement is a good training fit, but this is far from sufficient. Many (NG)RC models have extremely low training error, but fail completely during the prediction phase (\cref{fig:train-vs-predict-fail}). 
A stronger criterion germane to chaotic systems is that the predicted trajectory (beyond the training data) should reproduce the ``climate'' of the strange attractor, for example replicating the Lyapunov exponents \cite{pathak2017using}.
Here, we propose that the ability to accurately predict basins of attraction is another important test a model must pass before it can be trusted as a proxy of the underlying system. This applies as much to single-attractor systems as it does to multistable ones, as a model might produce spurious attractors not present in the original dynamics \cite{flynn2021multifunctionality}.

Here, we have shown that there exist commonly-studied systems for which basin prediction presents a steep challenge to leading RC frameworks. In standard RC, the model must be warmed up by an overwhelming majority of the transient dynamics, essentially reaching the attractor before prediction can begin. In contrast, NGRC requires minimal warmup data but is critically sensitive to the choice of readout nonlinearity, with its ability to make basin predictions contingent on having the exact features in the underlying dynamics. Though these frameworks face very different challenges, each presents a ``catch-22'': the dynamics cannot be learned unless key information about the system is already known. 

The basin prediction problem poses distinct challenges from the problem of forecasting chaotic systems, a test (NG)RC has largely passed with flying colors \cite{jaeger2004harnessing,pathak2017using,pathak2018model,carroll2018using,rafayelyan2020large,fan2020long,kim2021teaching,gauthier2021next,patel2023using}. In the latter case, the ``climate'' of a strange attractor can still be accurately reproduced even after the short-term prediction has failed \cite{pathak2017using}. It is for this reason that---in the most commonly-used benchmark systems (Lorenz-63, Lorenz-96, Kuramoto–Sivashinsky, etc.)---the transients are often deemed uninteresting and discarded during training. But for multistable systems, to predict which attractor an initial condition will converge to, the transient dynamics are the whole story. 
Therefore, basin prediction can be even more challenging than forecasting chaos. 
This is true even in the idealized setting considered here, wherein the attractors are fixed points, and the state of the system is fully observed without noise.
As such, we suggest that the magnetic pendulum and 
Kuramoto systems are ideal benchmarks for data-driven methods aiming to learn multistable nonlinear systems.

It has been established that both standard RC and NGRC are universal approximators, which in appropriate limits can achieve arbitrarily good fits to any system's dynamics \cite{gonon2019reservoir,hart2021echo}. But in practice, this is a rather weak guarantee. Unlike many other machine learning tasks, achieving a good fit to the flow of the real system [\cref{eq:discrete-dyn}] is only the first step; we must ultimately evolve the trained model as a dynamical system in its own right. This can invite a problem of stability, similar to the one faced by numerical integrators.
Even when the fit to a system's flow
is excellent, the autonomous dynamics of an (NG)RC model can be unstable, causing the prediction to diverge catastrophically from the true solution.
How to ensure the stability of a trained (NG)RC model in the general case is a major open problem \cite{lukovsevivcius2012practical}.

There are several exciting directions for future research that follow naturally from our results.
First, RC's ability to extract global information about a nonlinear system from local transient trajectories is one of its most powerful assets.
Currently, we lack a theory that characterizes conditions under which such extrapolations can be achieved by an RC model.
Second, several factors could contribute to the difficulty of basin prediction for RC, including the nonlinearity in the underlying equations, 
the geometric complexity of the basins, 
and the nature of the attractors themselves. 
Can we untangle the effects of these factors? 
Finally, although standard RC requires relatively long initialization data, it tends to show more robustness towards the choice of nonlinearity (i.e., the reservoir activation function) compared to NGRC.
Can we develop a new framework that combines standard RC's robustness with NGRC's efficiency and low data requirement?

RC is elegant, efficient, and powerful; but to usher in a new era of model-free learning of complex dynamical systems \cite{brunton2016discovering,weinan2017proposal,chen2018neural,li2020fourier,guimera2020bayesian,gilpin2020deep,karniadakis2021physics,nelsen2021random,belkin2021fit,levine2022framework,brunton2022modern}, it needs to solve the catch-22 created by its fragile dependence on readout nonlinearity (NGRC) or its reliance on long initialization data for every new initial condition (standard RC).

\begin{acknowledgments}
We thank D.~Gauthier, M.~Girvan, M.~Levine, and A.~Haji for insightful discussions.
Y.Z.\ acknowledges support from the Schmidt Science Fellowship and Omidyar Fellowship.
\end{acknowledgments}

\appendix

\section{Software implementation}
All simulations in this study were performed in Julia. For standard RC (Sections  \ref{sec:standard-rc-implementation}-\ref{sec:standard-rc-results}), we employ the \texttt{ReservoirComputing} package in concert with the \texttt{BayesianOptimization} package for hyperparameter optimization. For NGRC (Sections \ref{sec:ngrc-implementation}-\ref{sec:kuramoto}), we use a custom implementation as described in \cref{sec:ngrc-implementation}. Our source code is freely available online \footnote{Our source code can be found at \url{https://github.com/spcornelius/RCBasins}}. \\

\section{Numerical integration}
For the purpose of obtaining trajectories of the real system for training and validation, we use Julia's \texttt{DifferentialEquations} package to integrate all continuous equations of motion (\ref{eq:continuous-dyn}) using a $9$th-order integration scheme (Vern9), with absolute and relative error tolerances both set to $10^{-10}$. We stress that the hyperparameter $\Delta t$ has no relation to the numerical integration step size, which is determined adaptively to achieve the desired error tolerances. Instead, $\Delta t$ simply represents the timescale at which we seek to model the real dynamics via (NG)RC, and hence the resolution at which we sample the continuous trajectories to generate training and validation data. \\

\section{(Normalized) root-mean-square error}
\label{sec:nrmse}
Given an (NG)RC predicted trajectory $\mathbf{\tilde x}_t$ and a corresponding trajectory of the real system $\x_t$---each of length $N$---we calculate the root-mean-square error (RMSE) as
\begin{equation}
    \text{RMSE} = \sqrt{\frac{1}{N} \sum_t \lVert \x_t - \mathbf{\tilde x}_t \rVert^2},
\label{eq:rmse}
\end{equation}
where $\lVert \cdot \rVert$ denotes the Euclidean norm. To obtain a normalized version of this (NRMSE)---which we use as part of the objective function to optimize standard RC hyperparameters (\cref{sec:bayes_opt})---we first rescale each component of $\x_t$ and $\mathbf{\tilde x}_t$ by their range in the real system, \emph{e.g.},
\begin{equation}
    \x_{i,t} \rightarrow \frac{\x_{i,t}}{\x_{i,\max} - \x_{i, \min}},
\end{equation}
where the maximum ($\x_{i,\max}$) and minimum ($\x_{i, \min}$) for dimension $i=1,\ldots,n$ of the state space are calculated over the corresponding training data.

\section{Basin prediction}
\label{sec:basin_prediction}
We associate a given condition $\x_0$ with a basin of attraction by simulating the real (NGRC) dynamics for a total of $T$ time units ($\lceil T/\Delta t \rceil$ iterations). We then identify the closest stable fixed point at the end of the trajectory. In the magnetic pendulum, this is taken as the closest magnet. In the Kuramoto system, we calculate the winding number $|q|$ and use it to identify the corresponding twisted state. We use $T = 100$ for both systems, which is sufficient for all initial conditions under study to approach one of the stable fixed points.

\bibliography{bibli}

\clearpage

\newcommand\SupplementaryMaterials{%
  \xdef\presupfigures{\arabic{figure}}
  \xdef\presupsections{\arabic{section}}
  \xdef\presupequations{\arabic{equation}}
  \xdef\presuptables{\arabic{table}}
  \renewcommand\thefigure{S\fpeval{\arabic{figure}-\presupfigures}}
  \renewcommand\thesection{S\fpeval{\arabic{section}-\presupsections}}
  \renewcommand\theequation{S\fpeval{\arabic{equation}-\presupequations}}
  \renewcommand\thetable{S\fpeval{\arabic{table}-\presuptables}}
}

\SupplementaryMaterials

\clearpage
\onecolumngrid
\setcounter{page}{1}

\begin{center}
{\Large\bf Supplemental Material}\\[3mm]
{\large{Catch-22s of Reservoir Computing}}\\[1pt]
Yuanzhao Zhang and Sean P. Cornelius
\end{center}


\section{Supplemental Tables}

\begin{table*}[h]
\small
\begin{tabular}{cccc}
\toprule
Hyperparameter & Meaning & Lower Bound & Upper Bound \\
\midrule
$\rho$ & spectral radius of reservoir matrix ($\W$) & $10^{-3}$ & 1 \\
$s_x$ & input scaling (position) & $10^{-3}$ & 10 \\
$s_v$ & input scaling (velocity) & $10^{-3}$ & 10 \\
$s_b$ & bias scaling & $10^{-3}$ & 1 \\
$\alpha$ & leaky coefficient & $10^{-3}$ & 1 \\
\bottomrule
\end{tabular} 
\caption{\textbf{Optimizable hyperparameters in standard RC}. Each hyperparameter is optimized in logarithmic scale between the given bounds.}
\label{table:optimizable-rc-params}
\end{table*}

\begin{table*}[h]
\small
\begin{tabular}{ccccccc}
\toprule
\multicolumn{2}{c}{} & \multicolumn{5}{c}{Hyperparameter values} \\
\cmidrule(rl){3-7} 
Initial condition $(x_0, y_0)$ & Figures & $\rho$ & $s_x$ & $s_v$ & $s_b$ & $\alpha$ \\
\midrule
(-1.3, 0.75) & \ref{fig:err_vs_warmup1} and \ref{fig:err_vs_warmup_example1} & 
0.44077 & 5.5064 & 0.027882 & 1.0000 & 1.0000 \\
(1.0, -0.5) & \ref{fig:err_vs_warmup2} and \ref{fig:err_vs_warmup_example2} & 0.40633 & 5.0712 & 0.44366 & 1.0000 & 1.0000 \\
(1.75, 1.6) & \ref{fig:err_vs_warmup3} and \ref{fig:err_vs_warmup_example3} & 0.39391 & 2.9633 & 0.26557 & 1.0000 & 1.0000 \\
\midrule
\bottomrule
\end{tabular} 
\caption{\textbf{Values of optimized hyperparameters for standard RC}. We list the value of each hyperparameter after optimization to five significant figures. }
\label{table:rc-param-values}
\end{table*}

\section{Hyperparameter Optimization}
\label{sec:bayes_opt}

Given an initial condition $\mathbf{x}_0 = \left(x_0, y_0, \dot{x}_0, \dot{y}_0\right)^T$ of the magnetic pendulum system, we identify an optimal set of RC hyperparameters using Bayesian optimization. The goal here is to find the minimizer $\mathbf{p}^*$ of a (noisy) function $\mathcal{F}(\mathbf{p})$, i.e.,
\begin{equation}
    \mathbf{p}^* = \underset{\mathbf{p}}{\arg\min}\;\; \mathcal{F}(\mathbf{p}).
\end{equation}
In our setting, $\mathbf{p} = \left(\rho, s_x, s_v, s_b, \alpha\right)^T$ is a vector of our optimizable hyperparameters, and $\mathcal{F}$ is a scalar objective function measuring the error between the real system and a trained RC model generated with those hyperparameters. Typically, this objective function incorporates the NRMSE (\cref{sec:nrmse}) between the real and RC-predicted trajectories \cite{griffith2019forecasting}. But what is the best choice?

We found that the NRMSE during training is a poor optimization objective. In the magnetic pendulum, the resulting RC dynamics tend to blow up during the subsequent autonomous prediction, rather than staying near the fixed point of the real system. Accordingly, we use an objective function that incorporates both training \emph{and} validation NRMSE. Specifically, for a given set of hyperparameters $\*p$, we generate one random RC model and train it to the first $N_\text{train} = 4000$ steps of the real trajectory starting from $\mathbf{x}_0$. This yields a training NRMSE $\varepsilon_\text{train}$. We then simulate the trained RC model for an additional $N_\text{validation}$ time steps, picking up where the training left off. This yields a validation NRMSE, $\varepsilon_\text{validation}$. We then calculate $\mathcal{F}(\*p)$ as  
\begin{equation}
    \mathcal{F}(\*p) = \log (\varepsilon_\text{train}) + \log (\varepsilon_\text{validation}).
\end{equation}
 We find that this approach yields optimal RC models that have excellent training fits, but remain ``well-behaved'' (i.e., nearly stationary) beyond the training phase.

All hyperparameter optimization for standard RC was performed using the \texttt{BayesianOptimization} package in Julia. We model the landscape of $\mathcal{F}$ via Gaussian process regression to observed values of $\left(\mathbf{p}, \mathcal{F}\left(\mathbf{p}\right)\right)$. We employ the default squared-exponential (Gaussian) kernel, with tunable parameters corresponding to the standard deviation plus the length scale of each dimension of the hyperparameter space.  We first bootstrap the kernel (fit its parameters) using 200 random sets of hyperparameters $\mathbf{p}$ generated log-uniformly between the bounds in \cref{table:optimizable-rc-params} via Latin hypercube sampling. At every step of the process thereafter, we acquire a new candidate value of $\mathbf{p}$ via the commonly-used Expected Improvement strategy. We repeat this process for a total of 500 iterations, returning the observed minimizer of $\mathcal{F}(\*p)$. 
Every 50 iterations, we refit the kernel parameters via maximum {\it a posteriori} (MAP) estimation. To account for the stochasticity in $\mathcal{F}$ due to $\*W$, $\*W_\text{in}$, and $\*b$, we generate 10 realizations of the RC model at each candidate set of hyperparameters $\mathbf{p}$. Thus, over the course of the optimization, we evaluate $\mathcal{F}$ a total of 12000 times---2000 for the initial bootstrapping period, and an additional 10000 during the subsequent optimization.

\section{Supplemental Figures}

\begin{figure}[ht]
    \centering
    \includegraphics{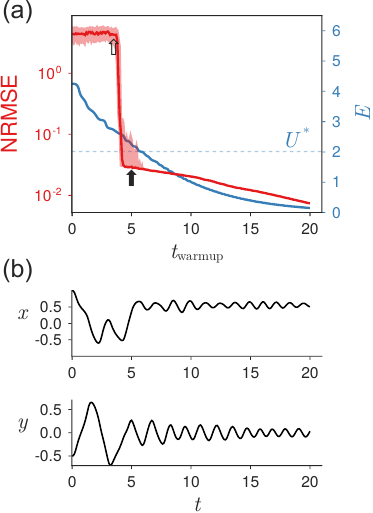}
    \caption{\textbf{Forecastability transition of standard RC.} Counterpart to \cref{fig:err_vs_warmup1} with the initial condition $(x_0, y_0) = (1.0, -0.5)$. The optimized RC hyperparameters for this initial condition are listed in \cref{table:rc-param-values}.
    }
    \label{fig:err_vs_warmup2}
\end{figure}

\begin{figure}[ht]
    \centering
    \includegraphics{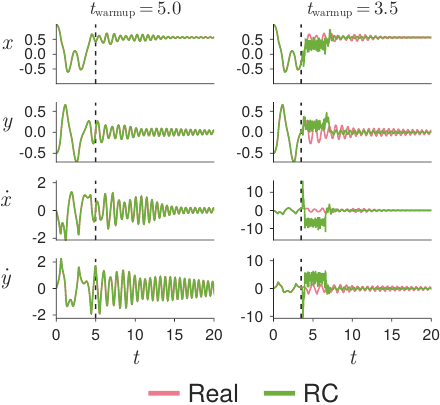}
    \caption{\textbf{Sensitivity of standard RC performance to warmup time.}
    Counterpart to \cref{fig:err_vs_warmup_example1} with the initial condition $(x_0, y_0) = (1.0, -0.5)$. The respective warmup times indicated by the dashed lines are the same as in \cref{fig:err_vs_warmup2}.
    }
    \label{fig:err_vs_warmup_example2}
\end{figure}

\begin{figure}[ht]
    \centering
    \includegraphics{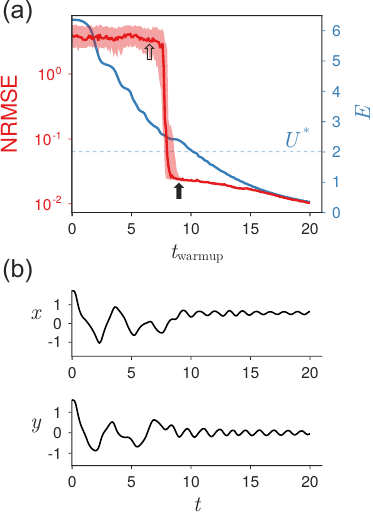}
    \caption{\textbf{Forecastability transition of standard RC.} Counterpart to \cref{fig:err_vs_warmup1} with the initial condition $(x_0, y_0) = (1.75, 1.6)$. The optimized RC hyperparameters for this initial condition are listed in \cref{table:rc-param-values}.
    }
    \label{fig:err_vs_warmup3}
\end{figure}

\begin{figure}[ht]
    \centering
    \includegraphics{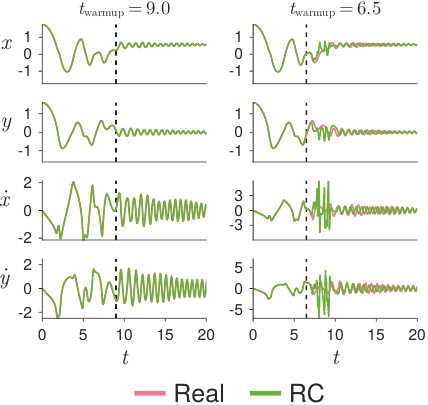}
    \caption{\textbf{Sensitivity of standard RC performance to warmup time.}
    Counterpart to \cref{fig:err_vs_warmup_example1} with the initial condition $(x_0, y_0) = (1.75, 1.6)$. The respective warmup times indicated by the dashed lines are the same as in \cref{fig:err_vs_warmup3}.
    }
    \label{fig:err_vs_warmup_example3}
\end{figure}

\begin{figure}[ht]
    \centering
    \includegraphics{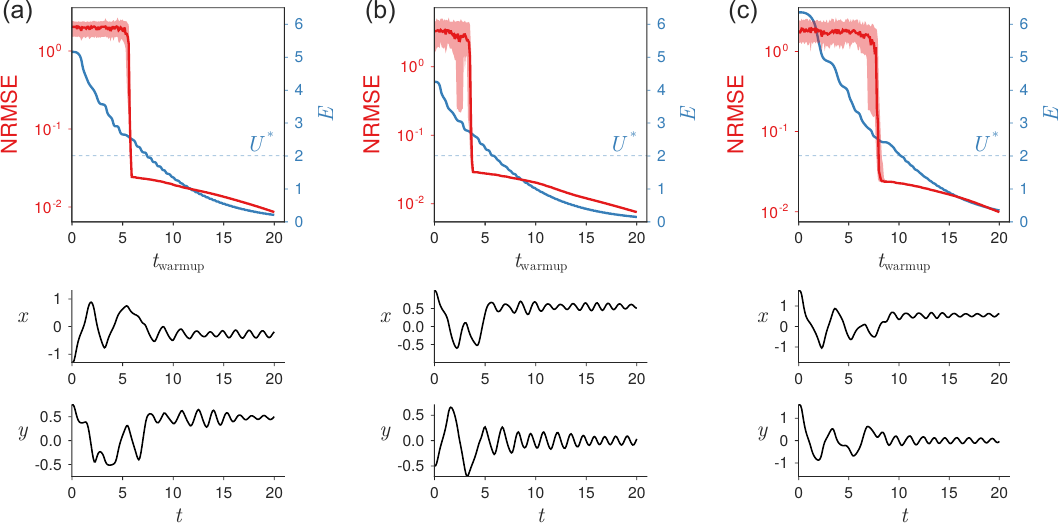}
    \caption{\textbf{Forecastability transition of standard RC.} Counterparts to \cref{fig:err_vs_warmup1}, \cref{fig:err_vs_warmup2}, and \cref{fig:err_vs_warmup3} using a larger reservoir size of $N_r = 600$.
    }
    \label{fig:err_vs_warmup_example_big}
\end{figure}

\begin{figure}[ht]
\centering
\includegraphics[width=.9\columnwidth]{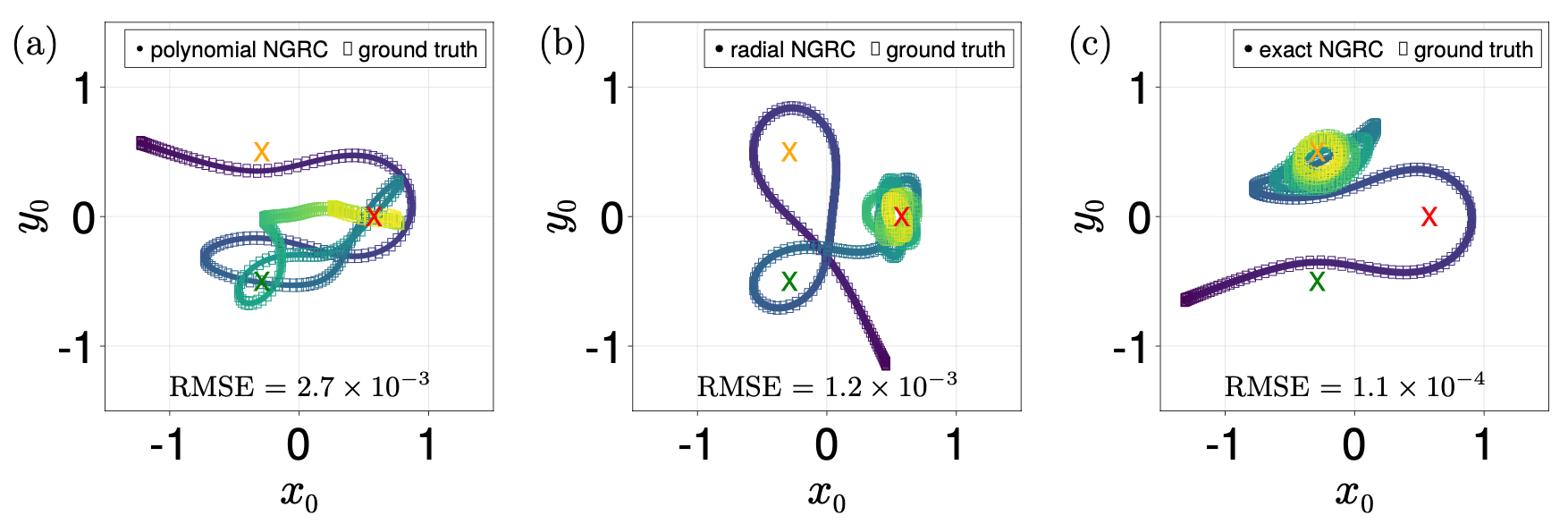}
\caption{
\textbf{NGRC models have excellent training fit for all readout nonlinearities tested.}
Each panel shows an NGRC model with a different readout nonlinearity: (a) polynomials with $d_{\max}=3$; (b) radial basis functions with $\NRBF=500$; (c) exact nonlinearity.
The trajectories are color-coded in time---they begin with dark purple points and end with bright green points.
The three fixed points are represented as crosses.
The root-mean-square error (RMSE) for each training trajectory is shown at the bottom of the panel.
The other hyperparameters used are $\Delta t = 0.01$, $\lambda = 1$, $k=2$, $\Ntraj = 100$, and $\Ntrain = 5000$.
}
\label{fig:training_fit}
\end{figure}

\begin{figure*}[h]
\centering
\includegraphics{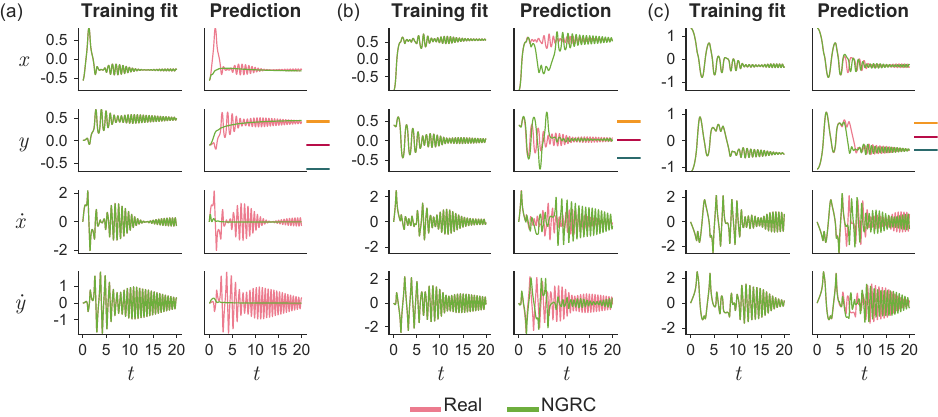}
\caption{
\textbf{NGRC models can fail to reproduce the correct transient dynamics even when the attractor is correctly predicted.} Counterpart to \cref{fig:train-vs-predict-fail}, showing examples of training initial conditions for which the NGRC predicted trajectory (green, right columns) goes to the correct attractor, but the transient dynamics differs markedly from the ground-truth (pink). 
All hyperparameters are the same as in \cref{fig:train-vs-predict-fail}.
}
\label{fig:train-vs-predict-success}
\end{figure*}

\begin{figure}[h]
\centering
\includegraphics[width=.5\columnwidth]{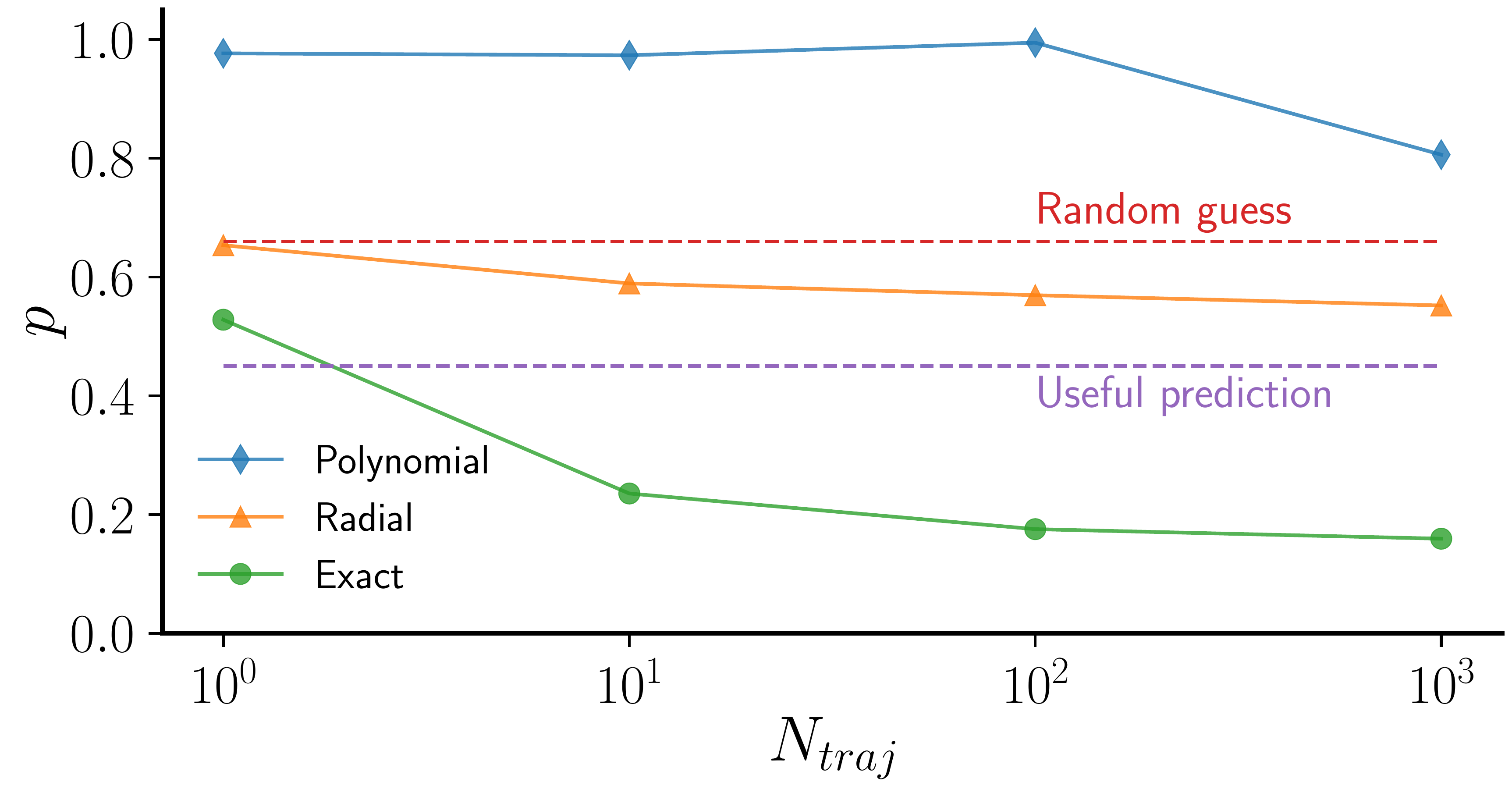}
\caption{
\textbf{Error rate $p$ as a function of the number of training trajectories $\Ntraj$ for NGRC models trained with polynomial, radial, and exact nonlinearity.} 
Each data point is obtained by averaging the error rate $p$ over $10$ independent realizations.
NGRC cannot produce useful predictions with polynomial nonlinearity ($d_{\max}=3$) or radial nonlinearity ($\NRBF=100$) no matter how many training trajectories are used.
With the exact nonlinearity from the magnetic pendulum equations, NGRC can make accurate predictions once trained on about $10$ trajectories.
Beyond this, more training trajectories yield only marginal increases in accuracy.
The other hyperparameters used here are $\Delta t = 0.01$, $k=2$, $\lambda = 1$, and $\Ntrain = 5000$.
}
\label{fig:error_n_trj}
\end{figure}

\begin{figure}[h]
\centering
\includegraphics[width=.5\columnwidth]{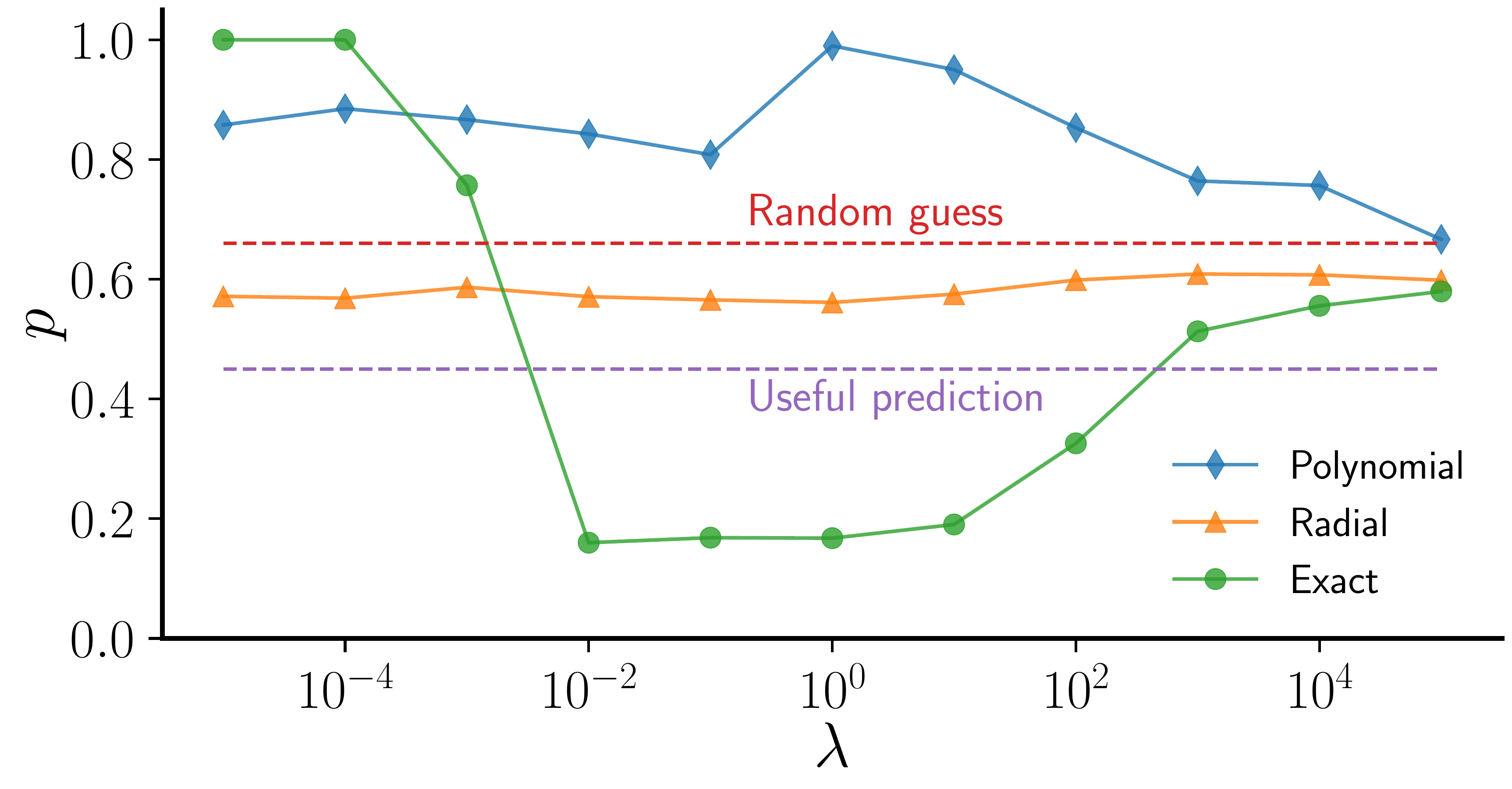}
\caption{
\textbf{Error rate $p$ as a function of the regularization coefficient $\lambda$ for NGRC models trained with polynomial, radial, and exact nonlinearity.} 
Each data point is obtained by averaging the error rate $p$ over $10$ independent realizations.
No choice of $\lambda$ can make NGRC produce useful predictions with polynomial nonlinearity ($d_{\max}=3$) or radial nonlinearity ($\NRBF=100$).
In contrast, exact nonlinearity can produce useful predictions for a wide range of $\lambda$ (between $10^{-2}$ and $10^2$).
The other hyperparameters used are $\Delta t = 0.01$, $k=2$, $\Ntraj = 100$, and $\Ntrain = 5000$.
}
\label{fig:error_lambda}
\end{figure}

\begin{figure}[h]
\centering
\includegraphics[width=.5\columnwidth]{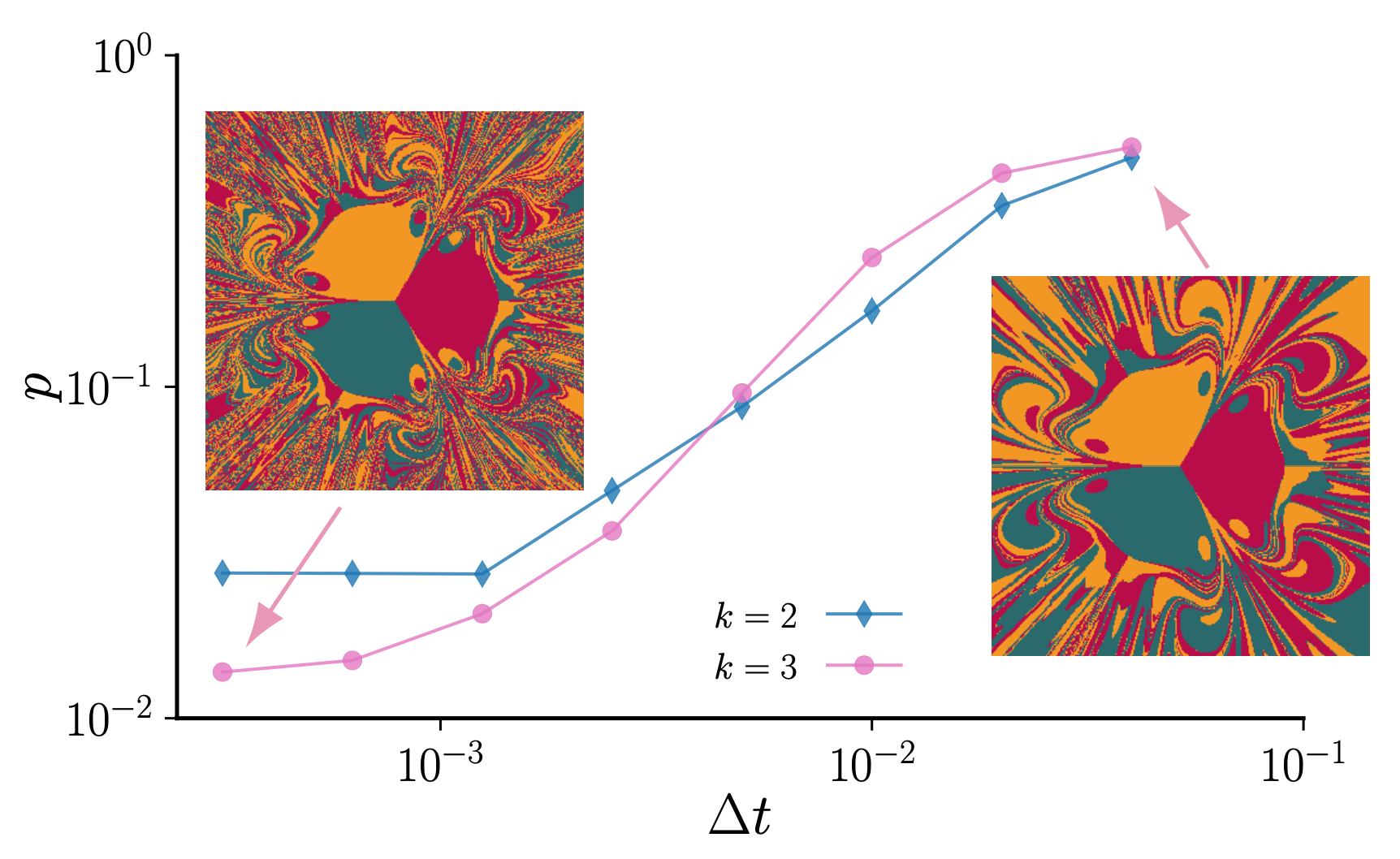}
\caption{
\textbf{Dependence of NGRC basin prediction accuracy on the time resolution $\Delta t$.}
Here, the NGRC models adopt the exact nonlinearity in the magnetic pendulum system.
Each data point is obtained by averaging the error rate $p$ over $10$ independent realizations (error bars are smaller than the size of the symbol).
The accuracy of the basin predictions can be significantly improved by taking smaller steps before it plateaus for $\Delta t$ below a certain threshold.
For $\Delta t=0.0003125$ (the leftmost points) and $k=3$, the NGRC model consistently achieves an accuracy around $98.6\%$.
Even for $\Delta t=0.04$ at the other end of the plot (right before NGRC becomes unstable and the solutions blow up), the features of the true basins are qualitatively preserved. 
Representative NGRC-predicted basins are shown for the two $\Delta t$ values discussed above.
The other hyperparameters used are $\lambda = 1$, $\Ntraj = 100$, and $\Ntrain = 20000$.
}
\label{fig:error_delta_t}
\end{figure}

\begin{figure}[h]
\centering
\includegraphics[width=1\columnwidth]{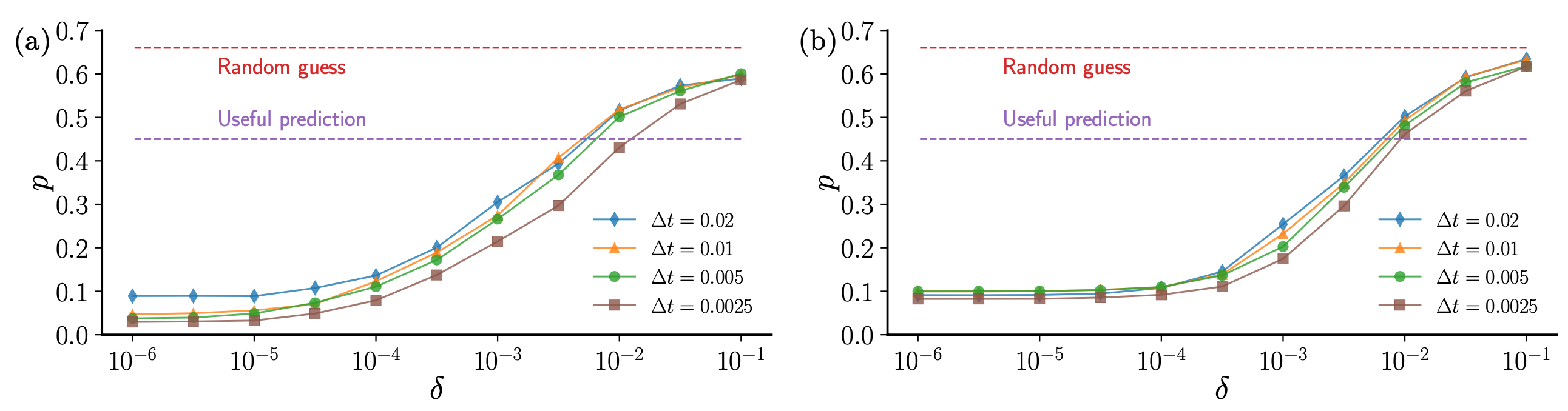}
\caption{
\textbf{NGRC basin prediction accuracy when using the exact nonlinearity but with small uncertainties.}
Same as \cref{fig:error_delta}, but with the height of the pendulum set to (a) $h=0.3$ and (b) $h=0.4$.
}
\label{fig:error_h}
\end{figure}

\begin{figure*}[h]
\centering
\includegraphics[width=.9\columnwidth]{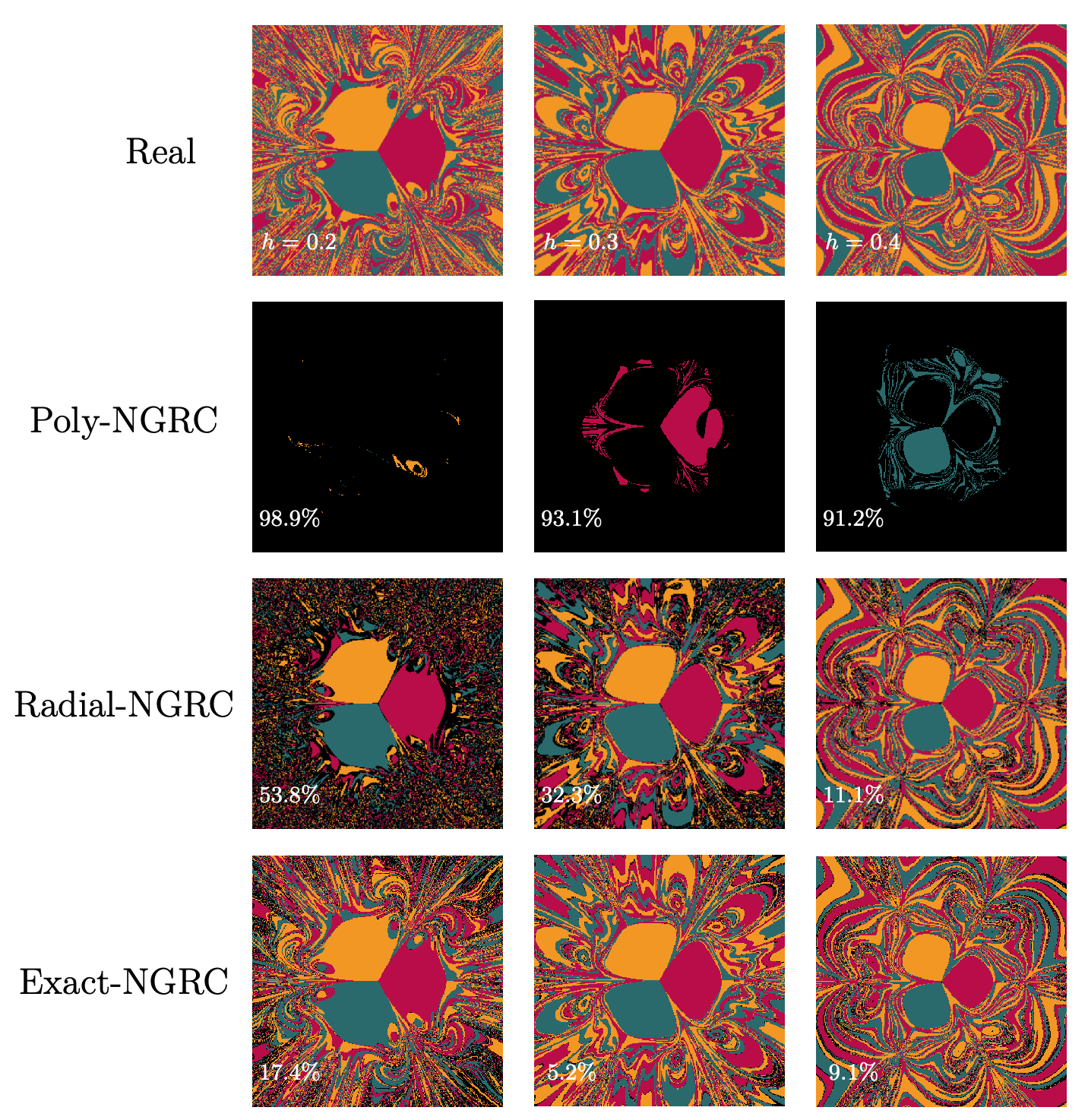}
\caption{
\textbf{Basin predictions generally become easier when the basins are less fractal.}
We show representative NGRC predictions for basins of the magnetic pendulum system with $h = 0.2$, $0.3$, and $0.4$.
As $h$ is increased, the basins become less fractal.
For the NGRC predictions, the error rate is marked in white and the wrong predictions are highlighted in black.
With polynomial nonlinearity ($d_{\max}=3$), NGRC predictions are worse than random guesses for all $h$ tested.
With radial nonlinearity ($\NRBF=500$), NGRC predictions become increasingly better as $h$ is increased.
With exact nonlinearity, NGRC predictions are consistently good, and the best accuracy is achieved at $h=0.3$ in this particular case.
The other hyperparameters used are $\Delta t = 0.01$, $\lambda = 1$, $k=2$, $\Ntraj = 100$, and $\Ntrain = 5000$.
}
\label{fig:h_dependence}
\end{figure*}

\end{document}